\documentclass{article}

\PassOptionsToPackage{numbers, compress}{natbib}
\usepackage[preprint]{neurips_2026}

\usepackage[utf8]{inputenc}
\usepackage[T1]{fontenc}
\usepackage{hyperref}
\usepackage{url}
\usepackage{booktabs}
\usepackage{amsfonts}
\usepackage{amsmath,amssymb}
\usepackage{nicefrac}
\usepackage{microtype}
\usepackage{algorithm}
\usepackage{algorithmic}
\usepackage{todonotes}
\usepackage{cleveref}
\usepackage{float}
\usepackage{sidecap}
\usepackage{tikz}
\usetikzlibrary{arrows.meta, positioning, calc, shapes.geometric,
               backgrounds, fit, decorations.pathreplacing}
\usepackage{xcolor}

\definecolor{stateblue}{HTML}{3B6EA5}
\definecolor{statefill}{HTML}{EBF1F8}
\definecolor{navteal}{HTML}{2A8C82}
\definecolor{navfill}{HTML}{E6F5F3}
\definecolor{latticeop}{HTML}{C06020}
\definecolor{latticefill}{HTML}{FDF0E6}
\definecolor{rewardgreen}{HTML}{3A7D44}
\definecolor{rewardfill}{HTML}{EAF5EB}
\definecolor{obspurple}{HTML}{6B4C9A}
\definecolor{obsfill}{HTML}{F0EBF5}
\definecolor{termred}{HTML}{B03A2E}
\definecolor{termfill}{HTML}{FBEAE8}
\definecolor{notegray}{HTML}{6C6C6C}
\definecolor{arrowgray}{HTML}{5A5A5A}
\definecolor{loopblue}{HTML}{5B8DB8}
\definecolor{compviolet}{HTML}{7B5EA7}
\definecolor{compfill}{HTML}{F3EFF8}
\definecolor{initgold}{HTML}{B58900}
\definecolor{initfill}{HTML}{FDF6E3}
\definecolor{bonusgreen}{HTML}{2E7D32}

\usetikzlibrary{positioning, fit, backgrounds, calc}
\usepackage{amsmath, amssymb}

\definecolor{leafamber}{HTML}{D97706}      
\definecolor{leaffill}{HTML}{FEF3C7}       

\definecolor{actteal}{HTML}{0D9488}        

\definecolor{netviolet}{HTML}{7C3AED}      
\definecolor{netfill}{HTML}{EDE9FE}        

\definecolor{backred}{HTML}{DC2626}        

\definecolor{edgegray}{HTML}{6B7280}       
\definecolor{annotgray}{HTML}{9CA3AF}      

\definecolor{headgreen}{HTML}{16A34A}      
\definecolor{headfill}{HTML}{DCFCE7}       

\definecolor{obspurple}{HTML}{9333EA}      
\definecolor{obsfill}{HTML}{F3E8FF}        

\newcommand{\circled}[1]{\tikz[baseline=(char.base)]{\node[shape=circle,draw=stateblue,fill=statefill,inner sep=1.5pt,font=\sffamily\bfseries\small] (char) {#1};}}

\definecolor{entlow}{HTML}{16A34A}
\definecolor{enthigh}{HTML}{DC2626}
\definecolor{horizblue}{HTML}{3B82F6}
\definecolor{horizfill}{HTML}{EFF6FF}
\definecolor{stopred}{HTML}{EF4444}

\newcommand{\Star}{{\texttt{Delta-Star}}}

\title{Discovering Lattice Reduction Strategies via Self-Play}

\author{%
  Mohamed Malhou \\
  FAIR, Meta Superintelligence Labs\\
  \& Sorbonne Université CNRS, LIP6 \\
 Paris, France \\
  \texttt{mmalhou@meta.com} \\
\And
Ludovic Perret  \\
EPITA, EPITA Research Lab (LRE)\\
Le Kremlin-Bicêtre, France  \\
\texttt{ludovic.perret@epita.fr}
\And
Kristin Lauter \\
FAIR, Meta Superintelligence Labs\\
\texttt{klauter@meta.com} \\
}

\begin{document}

\maketitle

\begin{abstract}
  The Lenstra-Lenstra-Lov\'asz (LLL) algorithm is a seminal contribution to computer science used for lattice basis reduction, yet its polynomial-time outputs produce bases that are far from optimal as the dimension grows.  We show that deep reinforcement learning can discover strictly superior, generalizable reduction strategies by interacting with the primitive action space of LLL. We formulate lattice reduction as a single-player Markov Decision Process (MDP) and train a deep residual network using an AlphaZero-style self-play pipeline augmented with adaptive-horizon MCTS (Monte Carlo Tree Search), which couples multi-step network predictions with an entropy-gated expansion mechanism. The resulting policy, \Star{}, is trained exclusively on small $8$-dimensional $q$-ary lattices and requires fewer primitive row operations than LLL. Crucially, it generalizes zero-shot to unseen moduli and higher dimensions up to $n=32$ without retraining.
\end{abstract}

\section{Introduction}
\label{sec:introduction}
Lattice reduction is a central problem in computational mathematics, underpinning modern algorithmic number theory, optimization, and cryptography. The security of modern lattice-based cryptographic schemes---including those recently standardized by NIST for post-quantum encryption \cite{nist2022finalists}---rests on the presumed hardness of finding short non-zero vectors in high-dimensional lattices \cite{ajtai1998shortest, micciancio2001complexity, Reg05}. Understanding the practical limits of lattice reduction algorithms is therefore critical for assessing the concrete security of these cryptosystems \cite{gama2008predicting}.

The workhorse of practical lattice reduction is the Lenstra-Lenstra-Lov\'asz (LLL) algorithm \cite{lenstra1982factoring}. LLL is a polynomial-time algorithm that produces a basis with guaranteed bounds on vector lengths. It operates through two local, primitive actions: \emph{size reduction}, which orthogonalizes a vector against its predecessors, and \emph{adjacent swaps}, which reorder the basis. The decision to swap is governed by the Lov\'asz condition, a handcrafted heuristic that ensures a monotonic decrease in a specific potential function.

While the Lov\'asz condition is sufficient to guarantee polynomial runtime, it is not necessarily optimal for achieving the strongest reduction in practice. The average-case behavior of LLL is notoriously difficult to analyze and often yields better bases than the worst-case bounds suggest \cite{gama2008predicting}. This gap raises a natural question: \emph{Can we discover a better sequencing policy for LLL's primitive operations that achieves stronger reduction than the classical algorithm?}

Recent work has shown that deep reinforcement learning (RL) combined with Monte Carlo Tree Search (MCTS) \cite{coulom2006efficient, kocsis2006bandit} can discover algorithms that outperform human-designed heuristics. Starting from complex board games \cite{silver2016mastering, silver2018general, schrittwieser2020muzero}, this approach has been adapted to discover new algorithms for matrix multiplication \cite{fawzi2022discovering}, sorting \cite{mankowitz2023alphadev}, combinatorial problems \cite{romera2024funsearch}, and formal mathematical reasoning \cite{hubert2025alphaproof}. These results demonstrate that when a mathematical problem is framed as a single-player game \cite{schadd2012single}, an RL agent can explore the space of possible algorithms and potentially identify strategies superior to those designed by hand. 

In parallel, neural methods have been explored directly for lattice reduction. \cite{neurallattice2023} proposed a self-supervised equivariant network that outputs factorized unimodular matrices, achieving performance comparable to LLL on benchmarks up to $n=8$ without explicit search. However, scaling such end-to-end approaches to higher dimensions remains challenging.

In this work, we apply the RL+MCTS paradigm to the core mechanics of lattice reduction. We formulate the lattice reduction process as a single-player Markov Decision Process (MDP), restricting our agent to the exact action space of the classical LLL algorithm: cursor movement, size reduction, and adjacent swaps. We train a deep residual network \cite{he2016resnet} using AlphaZero-style \cite{silver2018general} self-play and a novel adaptive-horizon MCTS to discover a data-driven policy that replaces the rigid Lov\'asz condition.

Our contributions are:
\begin{enumerate}
    \item \textbf{An RL Formulation for Lattice Reduction.} We define an MDP for lattice reduction that matches the local action space of LLL, enabling a direct comparison of sequencing policies without confounding factors from richer action spaces (such as exact Shortest-Vector-Problem (SVP) oracles \cite{gama2010extreme}).
    \item \textbf{Adaptive-Horizon MCTS.} We introduce an entropy-gated multi-step expansion mechanism for MCTS that uses the Shannon entropy of predicted policies to allow the search tree to skip deterministic continuations and amortize inference calls.
    \item \textbf{Beating LLL on its Own Terms.} Our learned policy achieves a significantly better root Hermite factor and orthogonality defect than the classical LLL algorithm across various dimensions, using only LLL's primitive operations.
    \item \textbf{Zero-Shot Generalization.} By using a $q$-normalized observation tensor and Resnet architecture with adaptive pooling on output features, our model---trained exclusively on small $8$-dimensional $q=251$-ary lattices---generalizes zero-shot to dimensions up to $n=32$ and unseen $q$ ($\sim 20-5000$).
\end{enumerate}

\section{Background}
\label{sec:background}
This section reviews the mathematical foundations of lattice reduction, with a focus on the LLL algorithm and the gap between its convergence guarantee and the quality of reduction it achieves in practice.

\subsection{Euclidean Lattices and Basis Quality}
\label{sec:metrics}
An $n$-dimensional lattice $\mathcal{L}$ is a discrete additive subgroup of $\mathbb{R}^n$ generated by $n$ linearly independent basis vectors $\mathbf{B} = \{\mathbf{b}_1, \dots, \mathbf{b}_n\}$:
$$\mathcal{L}(\mathbf{B}) = \left\{ \sum_{i=1}^n z_i \mathbf{b}_i : z_i \in \mathbb{Z} \right\}.$$
A lattice admits infinitely many bases: $\mathbf{B}_1$ and $\mathbf{B}_2$ generate the same lattice if and only if $\mathbf{B}_1 = \mathbf{B}_2 \mathbf{U}$ for some unimodular matrix $\mathbf{U} \in \mathbb{Z}^{n \times n}$ ($\det(\mathbf{U}) = \pm 1$). The volume $\det(\mathcal{L}) = |\det(\mathbf{B})|$ is a basis-invariant quantity.

\textbf{$q$-ary Lattices.} An integer lattice $\mathcal{L} \subseteq \mathbb{Z}^m$ is $q$-ary for a modulus $q \ge 2$ if $q\mathbb{Z}^m \subseteq \mathcal{L} \subseteq \mathbb{Z}^m$, so that membership depends only on the coordinates modulo $q$. For a matrix $\mathbf{A} \in \mathbb{Z}_q^{n \times m}$, the primal $q$-ary lattice is defined as
\begin{equation}
    \Lambda_q(\mathbf{A}) = \{ \mathbf{y} \in \mathbb{Z}^m : \exists \mathbf{s} \in \mathbb{Z}^n,\ \mathbf{s} \cdot \mathbf{A} \equiv \mathbf{y} \pmod{q} \}.
\end{equation}
To apply lattice reduction algorithms such as LLL or BKZ, we require an explicit full-rank basis for $\Lambda_q(\mathbf{A})$. We embed the lattice into $\mathbb{Z}^{n+m}$ using the following row-wise basis matrix:
\begin{equation}
    \mathbf{B} = \begin{bmatrix}
    q\mathbf{I}_m & \mathbf{0} \\
    \mathbf{A} & \mathbf{I}_n
    \end{bmatrix} \in \mathbb{Z}^{(n+m) \times (n+m)}.
\end{equation}
The rows of $\mathbf{B}$ generate $\Lambda_q(\mathbf{A})$ embedded in $\mathbb{Z}^{n+m}$, and the construction ensures $q\mathbb{Z}^{n+m} \subseteq \mathcal{L}(\mathbf{B})$.

\textbf{Basis quality.} The standard metric for evaluating a reduced basis is the \emph{root Hermite factor}:
\begin{equation}
    \delta_0 = \left( \frac{\|\mathbf{b}_1\|}{\det(\mathcal{L})^{1/n}} \right)^{1/n},
\end{equation}
which measures how short the first basis vector is relative to the lattice volume. A smaller $\delta_0$ indicates a stronger reduction. The global quality of the entire basis is captured by the \emph{orthogonality defect} $\delta(\mathbf{B}) = \prod_{i} \|\mathbf{b}_i\| / \det(\mathcal{L})$, which equals $1$ if and only if the basis is orthogonal (Hadamard's inequality \cite{hadamard1893resolution}). Additional background is provided in Appendix~\ref{sec:appendix_lattice_background}.

\subsection{Gram-Schmidt Orthogonalization}

To analyze the geometry of a lattice basis, it is standard to compute its Gram-Schmidt orthogonalization (GSO). The GSO vectors $\widetilde{\mathbf{B}} = \{\widetilde{\mathbf{b}}_1, \dots, \widetilde{\mathbf{b}}_n\}$ are defined iteratively: $\widetilde{\mathbf{b}}_1 = \mathbf{b}_1$, and for $j > 1$,
\begin{equation}
    \widetilde{\mathbf{b}}_j = \mathbf{b}_j - \sum_{i=1}^{j-1} \mu_{i,j} \widetilde{\mathbf{b}}_i, \quad \text{where} \quad \mu_{i,j} = \frac{\langle \mathbf{b}_j, \widetilde{\mathbf{b}}_i \rangle}{\langle \widetilde{\mathbf{b}}_i, \widetilde{\mathbf{b}}_i \rangle}.
\end{equation}
The GSO vectors are mutually orthogonal and span the same subspaces as the original basis vectors.

\subsection{The LLL Algorithm}
\label{sec:lll}

The Lenstra-Lenstra-Lov\'asz (LLL) algorithm \cite{lenstra1982factoring} operates through two local, primitive actions: \emph{size reduction}, which ensures the Gram-Schmidt coefficients satisfy $|\mu_{i,j}| \le 1/2$, and \emph{adjacent swaps}, which reorder basis vectors. The decision to swap is governed by the Lov\'asz condition: for a parameter $\delta \in (\frac{1}{4}, 1]$,
\begin{equation}
    \delta \|\widetilde{\mathbf{b}}_i\|^2 \le \|\mu_{i,i+1} \widetilde{\mathbf{b}}_i + \widetilde{\mathbf{b}}_{i+1}\|^2.
\end{equation}
When this condition is violated, the algorithm swaps rows $i$ and $i+1$ and decrements the cursor; otherwise it advances. The full pseudocode is given in Algorithm~\ref{alg:lll}.

\begin{algorithm}[h]
\footnotesize 
\caption{The LLL Algorithm}
\label{alg:lll}
\begin{algorithmic}[1]
\REQUIRE Basis $\mathbf{B} = \{\mathbf{b}_1, \dots, \mathbf{b}_n\}$, parameter $\delta \in (\frac{1}{4}, 1]$
\STATE Compute GSO vectors $\widetilde{\mathbf{B}}$ \& coeffs $\mu_{i,j}$; set $k \leftarrow 2$
\WHILE{$k \le n$}
    \FOR{$j = k-1, \dots, 1$}
        \IF{$|\mu_{j,k}| > 1/2$}
            \STATE $\mathbf{b}_k \leftarrow \mathbf{b}_k - \lfloor \mu_{j,k} \rceil \mathbf{b}_j$; update $\mu_{i,k}$ ($i \le j$) \COMMENT{\textsc{SizeReduce}}
        \ENDIF
    \ENDFOR
    \IF{$\delta \|\widetilde{\mathbf{b}}_{k-1}\|^2 > \|\mu_{k-1,k} \widetilde{\mathbf{b}}_{k-1} + \widetilde{\mathbf{b}}_k\|^2$}
        \STATE Swap $\mathbf{b}_k, \mathbf{b}_{k-1}$; update GSO \& coeffs; $k \leftarrow \max(2, k-1)$ \COMMENT{\textsc{Swap}}
    \ELSE
        \STATE $k \leftarrow k + 1$
    \ENDIF
\ENDWHILE
\RETURN LLL-reduced basis $\mathbf{B}$
\end{algorithmic}
\end{algorithm}

Termination is guaranteed by a potential function $\Phi(\mathbf{B}) = \prod_{i=1}^{n} \|\widetilde{\mathbf{b}}_i\|^{n-i+1}$ that strictly decreases with every swap. The Lov\'asz condition is a sufficient condition under which a swap reduces $\Phi$, but it is not necessarily the optimal sequencing policy for achieving the strongest reduction.

\textbf{BKZ.} Schnorr's Block Korkine-Zolotarev (BKZ) algorithm \cite{BKZ, schnorr1994lattice} generalizes the Lov\'asz condition to blocks of size $\beta \ge 2$, achieving smaller root Hermite factors at the cost of exponential complexity in $\beta$.

\section{Method for \Star{}}
\label{sec:method}
This section formalizes lattice reduction as a single-player Markov Decision Process (MDP), introduces our adaptive horizon MCTS mechanism for efficient tree search, and describes the neural network architecture and distributed training pipeline.

\subsection{Lattice Reduction as a Single-Player Game}
\label{sec:game}

We define the \texttt{latticeenv} environment as an episodic, deterministic MDP where an agent sequentially manipulates a lattice basis to improve its quality. The state representation mirrors the internal state of the classical LLL algorithm, tracking both the current basis and a row pointer (cursor) $k$.

\textbf{Basis Sampling and $q$-ary Lattices.} We train on $q$-ary lattices, the primary family of lattices used in cryptography. 

For a base dimension $n$, the agent manipulates a $2n \times 2n$ matrix. During training, entries of $\mathbf{A}$ are sampled uniformly from $[0, q-1]$.

\textbf{State Space $\mathcal{S}$.} At timestep $t$, the state $s_t$ consists of the basis matrix $\mathbf{B}_t$, its Gram-Schmidt orthogonalization $\mathbf{B}^*_t$ (see Appendix \ref{sec:appendix_lattice_background}), the Gram-Schmidt coefficients $\boldsymbol{\mu}_t$, the cursor position $k_t \in \{0, \dots, 2n-1\}$, and the number of remaining steps.

\textbf{Action Space $\mathcal{A}$.} The agent selects from four discrete actions, identical to the primitive operations available in LLL (\cref{sec:lll}):
\begin{enumerate}
    \item \textsc{MoveUp}: Decrement the cursor $k \leftarrow \max(1, k-1)$. $\; \texttt{\# not allowed to be in } k=0$
    \item \textsc{MoveDown}: Increment the cursor $k \leftarrow \min(2n-1, k+1)$.
    \item \textsc{Swap}: Exchange rows $\mathbf{b}_k$ and $\mathbf{b}_{k-1}$, then update $k \leftarrow \max(1, k-1)$.
    \item \textsc{SizeReduce}: Subtract integer multiples of previous rows from $\mathbf{b}_k$ whenever $|\mu_{k,j}| > 1/2$ for all $j < k$.
\end{enumerate}
Separating cursor movement from basis operations gives the agent complete flexibility to traverse the basis in any order, and limits the action set size to just $4$ actions.

\textbf{Reward Function.} Because the initial values of quality metrics (\cref{sec:metrics}) vary across random lattices, we define a normalized reward. Let $M(\mathbf{B})$ be a chosen metric (orthogonality defect $\delta$ or LLL potential $\Phi$, see appendix \ref{sec:appendix_lattice_background}). The step-wise reward is the normalized relative reduction in the log-metric:
\begin{equation}
    r_t^M = \frac{\log M(\mathbf{B}_{t-1}) - \log M(\mathbf{B}_t)}{\log M(\mathbf{B}_0) + \epsilon},
\end{equation}
where $\epsilon$ is a small constant for numerical stability. We define a hybrid reward parameterized by a potential weight $p \in [0, 1]$:
\begin{equation}
    r_t = (1 - p) \cdot r_t^{\delta} + p \cdot r_t^{\Phi},
\end{equation}
interpolating between the orthogonality defect and the potential function objectives. The rational behind this is that the size reduction changes the former while the swap action modifies the latter. At termination, a penalty proportional to the remaining defect is applied.

\textbf{Termination.} An episode terminates when the agent exhausts its maximum step horizon $T_{\max}$.

%

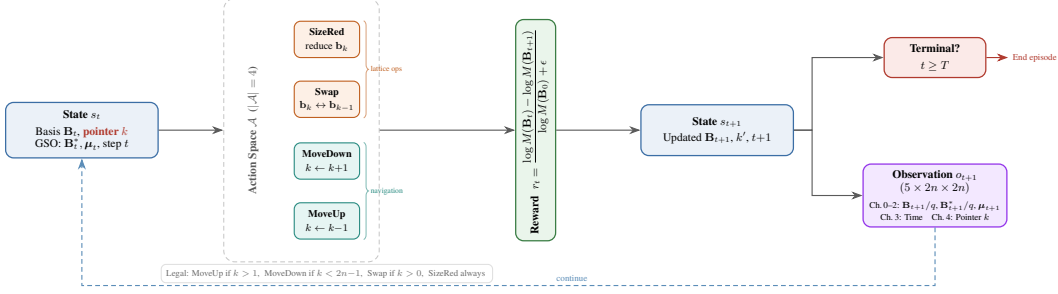
\begin{figure*}[htbp]
\centering
\resizebox{\textwidth}{!}{%
\begin{tikzpicture}[
    scale=0.8, transform shape, node distance=1.5cm and 2cm,
    statebox/.style={rectangle, rounded corners=5pt, draw=stateblue,
        fill=statefill, line width=0.8pt, minimum width=4.2cm,
        minimum height=1.4cm, font=\small, align=center, inner sep=6pt},
    actionbox/.style={rectangle, rounded corners=4pt, line width=0.7pt,
        minimum width=1.8cm, minimum height=1cm, font=\footnotesize,
        align=center, inner sep=4pt},
    rewardbox/.style={rectangle, rounded corners=5pt, draw=rewardgreen,
        fill=rewardfill, line width=0.8pt, minimum width=5.2cm,
        minimum height=1.1cm, font=\small, align=center, inner sep=5pt},
    obsbox/.style={rectangle, rounded corners=5pt, draw=obspurple,
        fill=obsfill, line width=0.8pt, minimum width=3.8cm,
        minimum height=1.7cm, font=\small, align=center, inner sep=5pt},
    termbox/.style={rectangle, rounded corners=5pt, draw=termred,
        fill=termfill, line width=0.8pt, minimum width=2.8cm,
        minimum height=1.1cm, font=\small, align=center, inner sep=5pt},
    arrow/.style={-{Stealth[length=2.5mm, width=1.8mm]},
        line width=0.7pt, draw=arrowgray},
    looparrow/.style={-{Stealth[length=2.5mm, width=1.8mm]},
        line width=0.7pt, draw=loopblue, densely dashed},
]

\node[statebox] (state) at (0, 0) {
    \textbf{State $s_t$}\\[3pt]
    Basis $\mathbf{B}_t$, \textcolor{termred}{\textbf{pointer $k$}}\\
    GSO: $\mathbf{B}^*_t, \boldsymbol{\mu}_t$, step $t$
};

\node[font=\small\bfseries, text=arrowgray, rotate=90, anchor=south]
    (actlabel) at (5.0, 0) {Action Space $\mathcal{A}$ \;($|\mathcal{A}|=4$)};

\node[actionbox, draw=navteal, fill=navfill] (moveup)
    at (6.75, -2.50) {\textbf{MoveUp}\\[1pt] $k \leftarrow k{-}1$};

\node[actionbox, draw=navteal, fill=navfill] (movedown)
    at (6.75, -0.85) {\textbf{MoveDown}\\[1pt] $k \leftarrow k{+}1$};

\node[actionbox, draw=latticeop, fill=latticefill] (swap)
    at (6.75, 0.85) {\textbf{Swap}\\[1pt] $\mathbf{b}_k \leftrightarrow \mathbf{b}_{k-1}$};

\node[actionbox, draw=latticeop, fill=latticefill] (sizered)
    at (6.75, 2.50) {\textbf{SizeRed}\\[1pt] reduce $\mathbf{b}_k$};

\begin{scope}[on background layer]
    \node[rectangle, rounded corners=7pt, draw=arrowgray!35, fill=white,
          densely dashed, line width=0.6pt, inner sep=12pt, inner ysep=14pt,
          fit=(moveup)(movedown)(swap)(sizered)(actlabel)] (actiongroup) {};
\end{scope}

\draw[decorate, decoration={brace, amplitude=3pt, mirror},
      draw=navteal!60, line width=0.5pt]
    ([xshift=0.1cm]moveup.south east) -- 
    node[right=2pt, font=\tiny, text=navteal] {navigation}
    ([xshift=0.1cm]movedown.north east);

\draw[decorate, decoration={brace, amplitude=3pt, mirror},
      draw=latticeop!60, line width=0.5pt]
    ([xshift=0.1cm]swap.south east) --
    node[right=2pt, font=\tiny, text=latticeop] {lattice ops}
    ([xshift=0.1cm]sizered.north east);

\node[rectangle, rounded corners=3pt, fill=white, draw=arrowgray!20,
      line width=0.4pt, font=\scriptsize, text=notegray, align=center,
      inner sep=4pt] (legal) at (6.75, -3.95) {
    Legal: MoveUp if $k>1$,\; MoveDown if $k<2n{-}1$,\;
    Swap if $k>0$,\; SizeRed always
};

\node[rewardbox, rotate=90] (reward) at (12.5, 0) {
    \textbf{Reward}\; $r_t = \dfrac{\log M(\mathbf{B}_{t}) - \log M(\mathbf{B}_{t+1})}{\log M(\mathbf{B}_0) + \epsilon}$
};

\node[statebox] (nextstate) at (17.5, 0) {
    \textbf{State $s_{t+1}$}\\[3pt]
    Updated $\mathbf{B}_{t+1}$, $k'$, $t{+}1$
};

\node[obsbox] (obs) at (23.5, -1.8) {
    \textbf{Observation $o_{t+1}$}\\
    $(5 \times 2n \times 2n)$\\[3pt]
    {\scriptsize Ch.\,0--2: $\mathbf{B}_{t+1}/q$, $\mathbf{B}^*_{t+1}/q$, $\boldsymbol{\mu}_{t+1}$}\\
    {\scriptsize Ch.\,3: Time\quad Ch.\,4: Pointer $k$}
};

\node[termbox] (terminal) at (23.5, 2.0) {
    \textbf{Terminal?}\\[3pt]
    $t \geq T$
};

\draw[arrow] (state.east) -- (actiongroup.west);
\draw[arrow] (actiongroup.east) -- (reward.north);
\draw[arrow] (reward.south) -- (nextstate.west);
\draw[arrow] (nextstate.east) -- ++(0.5,0) |- (obs.west);
\draw[arrow] (nextstate.east) -- ++(0.5,0) |- (terminal.west);

\draw[looparrow] (obs.south) -- ++(0,-1.6) -| (state.south);
\node[font=\scriptsize, text=loopblue, anchor=south, fill=white, inner sep=1pt]
    at (13.5, -4.2) {continue};

\draw[arrow, draw=termred] (terminal.east) -- ++(0.6,0)
    node[right, font=\scriptsize, text=termred] {End episode};

\end{tikzpicture}%
}
\caption{The \texttt{latticeenv} environment for LLL-style basis reduction. The agent controls a pointer $k$ indicating the current row, with four local actions. This mimics the sequential structure of classical LLL and enables learning better reduction strategies.}
\label{fig:latticelll-env}
\end{figure*}

\subsection{Adaptive Horizon Monte Carlo Tree Search}
\label{sec:adaptive_mcts}

We employ an AlphaZero-style self-play scheme \cite{silver2018general}, combining a dual-headed neural network $f_\theta(s) = (\mathbf{p}, v)$ with MCTS to iteratively improve the policy through self-play. The network is trained on data generated by MCTS to minimize the cross-entropy between its predicted policy $\mathbf{p}$ and the MCTS visit-count policy $\boldsymbol{\pi}$, and the mean squared error between its value estimate $v$ and the episode return $z$.

Standard AlphaZero MCTS expands one node per simulation: a leaf is evaluated by the network, its children are created, and the value is backed up. Our network produces multi-step horizon predictions---a policy $\mathbf{p}^{(k)}$ and value $v^{(k)}$ for each of $k = 0, \dots, H-1$ future steps---which allows expanding multiple nodes along the predicted greedy trajectory in a single evaluation, amortizing the inference computational cost.

However, the greedy path through horizon predictions is only reliable when the policy is concentrated. If a future step predicts a near-uniform distribution over actions, committing to the argmax is arbitrary and the value estimate is unreliable. We introduce an \emph{adaptive horizon} mechanism that uses the entropy of each predicted policy to detect transitions from forced sequences to genuine decision points, stopping expansion at the first uncertain step.

\textbf{Entropy-gated expansion.} Given logits $\log\mathbf{p}^{(k)}$, we compute its Shannon entropy: \newline $\mathbb{H}\bigl[\mathbf{p}^{(k)}\bigr] = -\sum_{a \in \mathcal{A}(s_k)} p_a^{(k)} \log_2 p_a^{(k)}.$

An entropy of $0$ indicates a single dominant action; $\mathbb{H} = 1.0$ implies two roughly equal options. When MCTS reaches a leaf, it queries the network once and expands nodes along the greedy path step-by-step. At each step $k$, if $\mathbb{H}[\mathbf{p}^{(k)}] \ge \tau$ for a predefined threshold $\tau$ (set to $0.6$ see Appendix \ref{sec:appendix_horizon_ablation}), the policy is too uncertain: the algorithm halts after expanding this node (making its children available for future PUCT selection). The value $v^{(k)}$ at the deepest expanded state is backpropagated through all newly created and visited horizon nodes and the selection path above them: $v_t = r_t + \lambda v_{t+1}$

This mechanism produces trees whose shape adapts to the problem structure. Forced sequences (e.g., navigation towards a specific row where $\mathbb{H} << 1$) are traversed automatically, preventing wasted simulations on deterministic continuations. However the starting behavior in early training is standard AlphaZero's MCTS \cite{silver2018general}. We provide a comprehensive ablation study of this mechanism's scaling dynamics in Appendix~\ref{sec:appendix_horizon_ablation}.

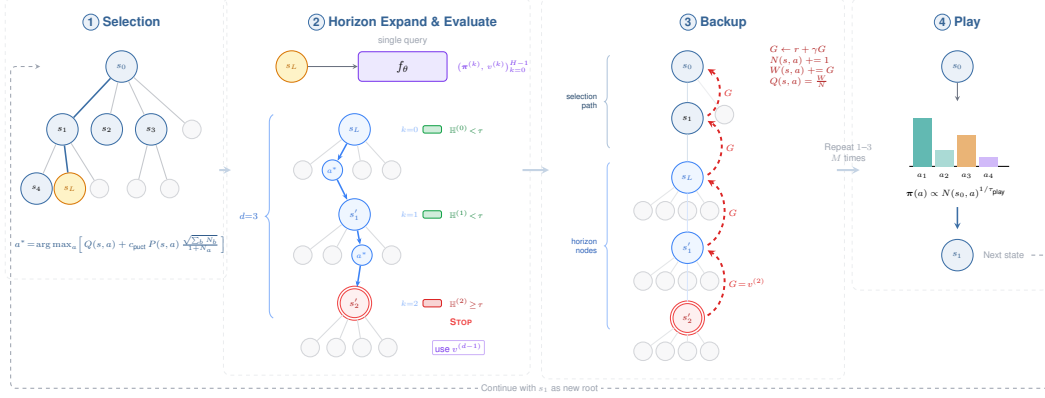
\begin{figure*}[t]
\centering
\resizebox{\textwidth}{!}{%
\begin{tikzpicture}[
    font=\sffamily,
    >=stealth,
    state/.style={circle, draw=stateblue, fill=statefill, thick,
                  minimum size=0.85cm, inner sep=0pt},
    leaf/.style={circle, draw=leafamber, fill=leaffill, thick,
                 minimum size=0.85cm, inner sep=0pt},
    horiznode/.style={circle, draw=horizblue, fill=horizfill, thick,
                      minimum size=0.85cm, inner sep=0pt},
    stopnode/.style={circle, draw=stopred, fill=stopred!8, thick,
                     minimum size=0.85cm, inner sep=0pt,
                     double, double distance=1.2pt},
    unvisited/.style={circle, draw=edgegray!30, fill=edgegray!5, thick,
                      minimum size=0.5cm, inner sep=0pt},
    greedychild/.style={circle, draw=horizblue, fill=horizfill, thick,
                        minimum size=0.55cm, inner sep=0pt},
    treeedge/.style={-, thick, draw=edgegray!40},
    seledge/.style={-, very thick, draw=stateblue},
    horizarrow/.style={-{Stealth[length=5pt]}, very thick, draw=horizblue},
    backarrow/.style={->, thick, dashed, draw=backred},
    panel/.style={rectangle, draw=annotgray!20, dashed,
                  rounded corners=5pt, inner sep=7pt},
    ptitle/.style={font=\sffamily\bfseries, text=stateblue!90!black},
    netbox/.style={rectangle, rounded corners=3pt, draw=netviolet,
                   fill=netfill, thick, align=center,
                   minimum height=0.65cm, minimum width=1.8cm},
    annot/.style={font=\sffamily\scriptsize, text=annotgray},
    entbar/.style={rectangle, rounded corners=1.5pt, minimum width=0.5cm,
                   minimum height=0.2cm, inner sep=0pt, thick},
]

\node[ptitle] at (2.2, 9.0) {\circled{1} Selection};

\node[state] (r) at (2.2, 7.8) {};
\node[font=\sffamily\scriptsize, text=stateblue!85!black] at (r) {$s_0$};

\node[state] (c1) at (0.6, 6.1) {};
\node[font=\sffamily\scriptsize] at (c1) {$s_1$};
\node[state] (c2) at (1.8, 6.1) {};
\node[font=\sffamily\scriptsize] at (c2) {$s_2$};
\node[state] (c3) at (3.0, 6.1) {};
\node[font=\sffamily\scriptsize] at (c3) {$s_3$};
\node[unvisited] (c4) at (4.1, 6.1) {};

\draw[seledge] (r) -- (c1);
\draw[treeedge] (r) -- (c2);
\draw[treeedge] (r) -- (c3);
\draw[treeedge] (r) -- (c4);

\node[state] (c1a) at (-0.1, 4.5) {};
\node[font=\sffamily\tiny] at (c1a) {$s_4$};
\node[leaf] (c1b) at (0.8, 4.5) {};
\node[font=\sffamily\scriptsize, text=leafamber!85!black] at (c1b) {$s_L$};
\node[unvisited] (c1c) at (1.6, 4.5) {};

\draw[treeedge] (c1) -- (c1a);
\draw[seledge] (c1) -- (c1b);
\draw[treeedge] (c1) -- (c1c);

\node[unvisited] (c3a) at (2.6, 4.5) {};
\node[unvisited] (c3b) at (3.5, 4.5) {};
\draw[treeedge] (c3) -- (c3a);
\draw[treeedge] (c3) -- (c3b);

\node[align=center, font=\sffamily\scriptsize, text=stateblue!85!black] at (2.2, 3.0) {
    $a^* \!=\! \arg\max_a \!\Big[\, Q(s,a) + c_{\text{puct}}\, P(s,a)\, \frac{\sqrt{\sum_b N_b}}{1{+}N_a} \,\Big]$
};

\node[ptitle] at (9.8, 9.0) {\circled{2} Horizon Expand \& Evaluate};

\node[netbox, minimum width=2.4cm, minimum height=0.8cm] (net) at (9.8, 7.8)
    {\textbf{$f_\theta$}};
\node[annot, above=1pt of net] {single query};

\node[leaf] (el) at (6.8, 7.8) {};
\node[font=\sffamily\scriptsize, text=leafamber!85!black] at (el) {$s_L$};
\draw[->, thick, draw=edgegray] (el.east) -- (net.west);

\node[font=\sffamily\tiny, text=netviolet, align=center, anchor=west] at (11.15, 7.8) {
    $\bigl(\boldsymbol{\pi}^{(k)},\, v^{(k)}\bigr)_{k=0}^{H-1}$
};

\node[horiznode] (h0) at (8.5, 6.1) {};
\node[font=\sffamily\scriptsize, text=horizblue!85!black] at (h0) {$s_L$};

\node[unvisited] (h0c1) at (7.1, 5.0) {};
\node[greedychild] (h0c2) at (7.9, 5.0) {};
\node[font=\sffamily\tiny, text=horizblue] at (h0c2) {$a^*$};
\node[unvisited] (h0c3) at (8.7, 5.0) {};
\node[unvisited] (h0c4) at (9.5, 5.0) {};
\draw[treeedge, draw=edgegray!25] (h0) -- (h0c1);
\draw[horizarrow] (h0) -- (h0c2);
\draw[treeedge, draw=edgegray!25] (h0) -- (h0c3);
\draw[treeedge, draw=edgegray!25] (h0) -- (h0c4);

\node[entbar, draw=entlow, fill=entlow!20] at (10.6, 6.1) {};
\node[font=\sffamily\tiny, text=entlow!80!black, anchor=west] at (10.95, 6.1)
    {$\mathbb{H}^{(0)} \!<\! \tau$};

\node[horiznode] (h1) at (8.5, 3.8) {};
\node[font=\sffamily\scriptsize, text=horizblue!85!black] at (h1) {$s'_1$};
\draw[horizarrow] (h0c2) -- (h1);

\node[unvisited] (h1c1) at (7.1, 2.7) {};
\node[unvisited] (h1c2) at (7.9, 2.7) {};
\node[greedychild] (h1c3) at (8.7, 2.7) {};
\node[font=\sffamily\tiny, text=horizblue] at (h1c3) {$a^*$};
\node[unvisited] (h1c4) at (9.5, 2.7) {};
\draw[treeedge, draw=edgegray!25] (h1) -- (h1c1);
\draw[treeedge, draw=edgegray!25] (h1) -- (h1c2);
\draw[horizarrow] (h1) -- (h1c3);
\draw[treeedge, draw=edgegray!25] (h1) -- (h1c4);

\node[entbar, draw=entlow, fill=entlow!20] at (10.6, 3.8) {};
\node[font=\sffamily\tiny, text=entlow!80!black, anchor=west] at (10.95, 3.8)
    {$\mathbb{H}^{(1)} \!<\! \tau$};

\node[stopnode] (h2) at (8.5, 1.4) {};
\node[font=\sffamily\scriptsize, text=enthigh!80!black] at (h2) {$s'_2$};
\draw[horizarrow] (h1c3) -- (h2);

\node[unvisited] (h2c1) at (7.3, 0.2) {};
\node[unvisited] (h2c2) at (8.0, 0.2) {};
\node[unvisited] (h2c3) at (8.7, 0.2) {};
\node[unvisited] (h2c4) at (9.4, 0.2) {};
\draw[treeedge, draw=edgegray!25] (h2) -- (h2c1);
\draw[treeedge, draw=edgegray!25] (h2) -- (h2c2);
\draw[treeedge, draw=edgegray!25] (h2) -- (h2c3);
\draw[treeedge, draw=edgegray!25] (h2) -- (h2c4);

\node[entbar, draw=enthigh, fill=enthigh!20] at (10.6, 1.4) {};
\node[font=\sffamily\tiny, text=enthigh!80!black, anchor=west] at (10.95, 1.4)
    {$\mathbb{H}^{(2)} \!\ge\! \tau$};
\node[font=\sffamily\scriptsize\bfseries, text=stopred, anchor=west] at (10.95, 0.9)
    {\textsc{Stop}};

\node[font=\sffamily\scriptsize, text=netviolet, fill=white, inner sep=2pt,
      rounded corners=1pt, draw=netviolet!40, anchor=west] at (10.6, 0.2)
    {use $v^{(d{-}1)}$};

\node[font=\sffamily\tiny, text=horizblue!60, anchor=east] at (10.35, 6.1) {$k{=}0$};
\node[font=\sffamily\tiny, text=horizblue!60, anchor=east] at (10.35, 3.8) {$k{=}1$};
\node[font=\sffamily\tiny, text=horizblue!60, anchor=east] at (10.35, 1.4) {$k{=}2$};

\draw[decorate, decoration={brace, amplitude=6pt, mirror}, thick, draw=horizblue!60]
    (6.3, 6.5) -- (6.3, 1.0)
    node[midway, left=8pt, font=\sffamily\scriptsize, text=horizblue!85!black, align=center]
    {$d{=}3$};

\node[ptitle] at (18.2, 9.0) {\circled{3} Backup};

\node[state] (br) at (17.5, 7.8) {};
\node[font=\sffamily\scriptsize, text=stateblue!85!black] at (br) {$s_0$};

\node[unvisited] (bsib1) at (18.5, 6.5) {};
\draw[treeedge, draw=edgegray!12] (br) -- (bsib1);

\node[state] (bc1) at (17.5, 6.4) {};
\node[font=\sffamily\scriptsize] at (bc1) {$s_1$};
\draw[treeedge, draw=stateblue!20] (br) -- (bc1);

\node[horiznode] (bh0) at (17.5, 4.8) {};
\node[font=\sffamily\scriptsize, text=horizblue!85!black] at (bh0) {$s_L$};
\draw[treeedge, draw=stateblue!20] (bc1) -- (bh0);

\node[unvisited] (bh0c1) at (16.3, 3.9) {};
\node[unvisited] (bh0c2) at (16.9, 3.9) {};
\node[unvisited] (bh0c3) at (17.5, 3.9) {};
\node[unvisited] (bh0c4) at (18.1, 3.9) {};
\draw[treeedge, draw=edgegray!18] (bh0) -- (bh0c1);
\draw[treeedge, draw=edgegray!18] (bh0) -- (bh0c2);
\draw[treeedge, draw=horizblue!18] (bh0) -- (bh0c3);
\draw[treeedge, draw=edgegray!18] (bh0) -- (bh0c4);

\node[horiznode] (bh1) at (17.5, 2.9) {};
\node[font=\sffamily\scriptsize, text=horizblue!85!black] at (bh1) {$s'_1$};
\draw[treeedge, draw=horizblue!20] (bh0) -- (bh1);

\node[unvisited] (bh1c1) at (16.3, 2.0) {};
\node[unvisited] (bh1c2) at (16.9, 2.0) {};
\node[unvisited] (bh1c3) at (17.5, 2.0) {};
\node[unvisited] (bh1c4) at (18.1, 2.0) {};
\draw[treeedge, draw=edgegray!18] (bh1) -- (bh1c1);
\draw[treeedge, draw=edgegray!18] (bh1) -- (bh1c2);
\draw[treeedge, draw=horizblue!18] (bh1) -- (bh1c3);
\draw[treeedge, draw=edgegray!18] (bh1) -- (bh1c4);

\node[stopnode] (bh2) at (17.5, 1.0) {};
\node[font=\sffamily\scriptsize, text=enthigh!80!black] at (bh2) {$s'_2$};
\draw[treeedge, draw=horizblue!20] (bh1) -- (bh2);

\node[unvisited] (bh2c1) at (16.3, 0.1) {};
\node[unvisited] (bh2c2) at (16.9, 0.1) {};
\node[unvisited] (bh2c3) at (17.5, 0.1) {};
\node[unvisited] (bh2c4) at (18.1, 0.1) {};
\draw[treeedge, draw=edgegray!18] (bh2) -- (bh2c1);
\draw[treeedge, draw=edgegray!18] (bh2) -- (bh2c2);
\draw[treeedge, draw=edgegray!18] (bh2) -- (bh2c3);
\draw[treeedge, draw=edgegray!18] (bh2) -- (bh2c4);

\draw[backarrow, line width=1.5pt]
    ([xshift=3pt, yshift=2pt]bh2.east) to[out=30, in=-30]
    node[font=\sffamily\scriptsize, right=2pt, text=backred, fill=white, inner sep=1pt]
    {$G \!=\! v^{(2)}$}
    ([xshift=3pt, yshift=-2pt]bh1.east);
\draw[backarrow, line width=1.5pt]
    ([xshift=3pt, yshift=2pt]bh1.east) to[out=30, in=-30]
    node[font=\sffamily\scriptsize, right=2pt, text=backred, fill=white, inner sep=1pt]
    {$G$}
    ([xshift=3pt, yshift=-2pt]bh0.east);
\draw[backarrow, line width=1.5pt]
    ([xshift=3pt, yshift=2pt]bh0.east) to[out=30, in=-30]
    node[font=\sffamily\scriptsize, right=2pt, text=backred, fill=white, inner sep=1pt]
    {$G$}
    ([xshift=3pt, yshift=-2pt]bc1.east);
\draw[backarrow, line width=1.5pt]
    ([xshift=3pt, yshift=2pt]bc1.east) to[out=30, in=-30]
    node[font=\sffamily\scriptsize, right=2pt, text=backred, fill=white, inner sep=1pt]
    {$G$}
    ([xshift=3pt, yshift=-2pt]br.east);

\node[align=left, font=\sffamily\scriptsize, text=backred!80!black] at (20.6, 7.8) {
    $G \leftarrow r + \gamma G$\\
    $N(s,a) \mathrel{+}= 1$\\
    $W(s,a) \mathrel{+}= G$\\
    $Q(s,a) = \frac{W}{N}$
};

\draw[decorate, decoration={brace, amplitude=4pt, mirror}, thick, draw=stateblue!40]
    (15.4, 8.1) -- (15.4, 5.6)
    node[midway, left=6pt, font=\sffamily\tiny, text=stateblue!65!black, align=right]
    {selection\\path};

\draw[decorate, decoration={brace, amplitude=4pt, mirror}, thick, draw=horizblue!50]
    (15.4, 5.2) -- (15.4, 0.7)
    node[midway, left=6pt, font=\sffamily\tiny, text=horizblue!75!black, align=right]
    {horizon\\nodes};

\node[ptitle] at (24.8, 9.0) {\circled{4} Play};

\node[state] (pr) at (24.8, 7.8) {};
\node[font=\sffamily\scriptsize, text=stateblue!85!black] at (pr) {$s_0$};

\draw[->, thick, draw=edgegray] (pr.south) -- ++(0, -0.5);

\begin{scope}[shift={(23.6, 5.1)}]
    \draw[thick, -] (-0.1,0) -- (2.6,0);
    \fill[actteal!65]    (0.0, 0) rectangle (0.5, 1.3);
    \fill[actteal!35]    (0.6, 0) rectangle (1.1, 0.45);
    \fill[leafamber!55]  (1.2, 0) rectangle (1.7, 0.85);
    \fill[netviolet!35]  (1.8, 0) rectangle (2.3, 0.25);
    \node[font=\sffamily\tiny, below=2pt] at (0.25, 0) {$a_1$};
    \node[font=\sffamily\tiny, below=2pt] at (0.85, 0) {$a_2$};
    \node[font=\sffamily\tiny, below=2pt] at (1.45, 0) {$a_3$};
    \node[font=\sffamily\tiny, below=2pt] at (2.05, 0) {$a_4$};
\end{scope}

\node[align=center, font=\sffamily\scriptsize] at (24.8, 4.4) {
    $\boldsymbol{\pi}(a) \propto N(s_0, a)^{1/\tau_{\text{play}}}$
};

\draw[->, very thick, draw=stateblue] (24.8, 4.0) -- (24.8, 3.4);
\node[state] (pnext) at (24.8, 2.7) {};
\node[font=\sffamily\scriptsize, text=stateblue!85!black] at (pnext) {$s_1$};
\node[annot, right=4pt of pnext] {Next state};

\begin{scope}[on background layer]
    \node[panel, fit={(-0.7, 2.3) (4.8, 9.4)}] {};
    \node[panel, fit={(5.2, -0.3) (13.0, 9.4)}] {};
    \node[panel, fit={(13.8, -0.4) (21.5, 9.4)}] {};
    \node[panel, fit={(22.2, 2.0) (27.0, 9.4)}] {};
\end{scope}

\draw[->, ultra thick, draw=stateblue!30, shorten >=4pt, shorten <=4pt]
    (4.7, 5.0) -- (5.3, 5.0);
\draw[->, ultra thick, draw=stateblue!30, shorten >=4pt, shorten <=4pt]
    (12.9, 5.0) -- (13.9, 5.0);
\draw[->, ultra thick, draw=stateblue!30, shorten >=4pt, shorten <=4pt]
    (21.4, 5.0) -- (22.3, 5.0)
    node[midway, above=3pt, font=\sffamily\scriptsize, text=annotgray, align=center]
    {Repeat $1{\text{--}}3$\\$M$ times};

\draw[->, thick, draw=annotgray, dashed, rounded corners=4pt]
    (26.8, 2.7) -- (27.5, 2.7) -- (27.5, -0.9) -- (-0.8, -0.9) -- (-0.8, 7.8) -- (0.0, 7.8);
\node[font=\sffamily\scriptsize, text=annotgray, fill=white, inner sep=2pt]
    at (13.5, -0.9) {Continue with $s_1$ as new root};

\end{tikzpicture}
}
\caption{Adaptive Horizon MCTS loop.
\textbf{(1)~Selection:} Starting from the root $s_0$, the tree is traversed by PUCT (Predictive Upper Confidence Bound for Trees \cite{Rosin2011,silver2016mastering}) to reach a leaf $s_L$.
\textbf{(2)~Horizon Expand \& Evaluate:} The network $f_\theta$ is queried once, producing horizon predictions $(\boldsymbol{\pi}^{(k)}, v^{(k)})_{k=0}^{H-1}$. Nodes are expanded along the greedy path; at each step the Shannon entropy $\mathbb{H}[\mathbf{p}^{(k)}]$ is checked against a threshold $\tau$. Low-entropy steps (confident policy) are traversed automatically; expansion halts at the first high-entropy step (uncertain decision point), yielding effective depth $d \le H$. All children at every expanded node are created, making them available for future PUCT selection.
\textbf{(3)~Backup:} The value $v^{(d-1)}$ from the deepest expanded state initializes the return $G$. This return is propagated back through the horizon chain and the original selection path, accumulating intermediate rewards $r$ with discount factor $\gamma$ ($G \leftarrow r + \gamma G$), updating $N$, $W$, and $Q$.
\textbf{(4)~Play:} After $M$ simulations the root visit counts determine the target policy $\boldsymbol{\pi}$.}
\label{fig:horizon_mcts_loop}
\end{figure*}

\subsection{Neural Network Architecture}
\label{sec:architecture}

The agent's policy and value estimates are parameterized by a deep Residual Network (ResNet) \cite{he2016resnet}. The architecture incorporates a lookback window for the input and a multi-step prediction horizon for the output.

\textbf{Observation Tensor.} The state $s_t$ is encoded into a 5 channel tensor ($5 \times 2n \times 2n$): \textbf{[0]} modulus-normalized basis $\mathbf{B} / q$; \textbf{[1]} modulus-normalized GSO vectors $\mathbf{B}^* / q$; \textbf{[2]} Gram-Schmidt coefficients $\boldsymbol{\mu}$; \textbf{[3]} normalized remaining time $(T_{\max} - t) / T_{\max}$; and \textbf{[4]} a one-hot cursor indicator where the $k$-th row is 1.

\textbf{Temporal Lookback.} The network receives a stacked history of the $W$ most recent observations along the channel dimension, resulting in an input tensor of shape $5W \times 2n \times 2n$.

\textbf{ResNet Backbone.} The input tensor is processed by an initial convolutional block (kernel size 3, padding 1) followed by Batch Normalization and ReLU activation, then passed through $D$ residual blocks with skip connections. We use a width of 256 filters and depth $D=10$.

\textbf{Horizon Prediction Heads.} We use adaptive pooling to collapse the spatial dimensions. The network predicts $H$ steps into the future simultaneously, splitting into two heads:
\begin{itemize}
    \item \textbf{Policy Head:} Outputs a tensor of shape $H \times |\mathcal{A}|$, representing action probabilities for the current state and $H-1$ subsequent states along the greedy path.
    \item \textbf{Value Head:} Outputs a tensor of shape $H$, representing expected returns for the current and future states.
\end{itemize}

\textbf{Loss Function.} We minimize a horizon-weighted loss:
\begin{equation}
    \mathcal{L}(\theta) = \frac{1}{\sum_k \lambda^k m_k} \sum_{k=0}^{H-1} \lambda^k \, m_k \Bigl[\ell_\text{CE}\bigl(\mathbf{p}^{(k)}, \hat{\boldsymbol{\pi}}^{(k)}\bigr) + c_v \, \ell_\text{MSE}\bigl(v^{(k)}, \hat{v}^{(k)}\bigr)\Bigr] + c\|\theta\|^2,
\end{equation}
where $\lambda \in (0, 1)$ is a horizon decay factor giving near-future predictions more weight, $m_k \in \{0, 1\}$ is a validity mask (zero when the trajectory has fewer than $k$ remaining steps), $\hat{\boldsymbol{\pi}}^{(k)}$ is the MCTS visit-count policy at step $t+k$, and $\hat{v}^{(k)}$ is the return from step $t+k$.

\subsection{Distributed Training Pipeline}
\label{sec:training}

Training requires generating millions of lattice reduction trajectories. We implement an asynchronous distributed system using the Ray framework \cite{Moritzetal2018}, dividing the workload across three actor types.

\textbf{Self-Play Workers (CPU).} Hundreds of parallel workers run the \texttt{latticeenv} environment. Each worker maintains a batch of active games and executes MCTS in parallel. Leaf nodes are queued for evaluation and sent as asynchronous requests to inference servers.

\textbf{Inference Servers (GPU).} Dedicated GPU servers process batched requests from the self-play workers. Requests are queued until a maximum batch size is reached or a timeout expires.

\textbf{The Learner (GPU).} A central Learner maintains the master weights, an Adam optimizer \cite{Kingma2015Adam}, and a trajectory-aware replay buffer. Workers send full trajectories. The Learner constructs target sequences of length $H$ for each sampled state and optimizes the loss function described above. The Learner periodically broadcasts updated weights to the inference servers.


\section{Experiments}
\label{sec:experiments}
\subsection{Experimental Setup}
\label{sec:setup}

We train the reinforcement learning agent, denoted \Star{}, on randomly generated $q$-ary lattices of base dimension $n=8$. As described in Section~\ref{sec:game}, the actual basis matrix is a $16 \times 16$ block matrix embedding the modulus $q=251$ and the identity. The agent is trained using the distributed pipeline of Section~\ref{sec:training} with the hybrid reward at potential weight $p=0.75$, selected based on the ablation study in Appendix~\ref{sec:appendix_reward_ablation}.

During evaluation, we deploy the trained policy stochastically with a low temperature $T = \frac{1}{2}$ and evaluate over 100 random lattice instances per condition. The primary metric is the root Hermite factor $\delta_0 = (\|\mathbf{b}_{\text{min}}\| / \det(\mathcal{L})^{1/n})^{1/n}$. We also track the orthogonality defect. Baselines are the classical LLL algorithm with $\delta_{LLL}=0.99$ and BKZ-$\beta$ with block size $\beta = \min(2n, 20)$ which is large enough for such small dimensions with a reasonable runtime. We track cumulative row operations as a proxy for algorithmic complexity. Each swap action counts as a single operation and navigation actions incur zero computational cost. Size-reduction computational cost is counted dynamically: one operation per index $k$ satisfying $|\mu_{j,k}| > \tfrac{1}{2}$, yielding between $0$ and $2n - 1$ operations per size-reduction step.

\subsection{Scaling to Large Dimensions: Approaching BKZ Quality}
\label{sec:scaling}

A key test for any learned combinatorial policy is zero-shot generalization: can a policy trained on small instances scale to larger, unseen dimensions? We evaluate \Star{}---trained exclusively on $n=8$---on lattices up to $n=32$ without further fine-tuning.

\begin{figure}[ht]
    \begin{minipage}[c]{0.45\textwidth}
        \centering
        \includegraphics[width=\textwidth]{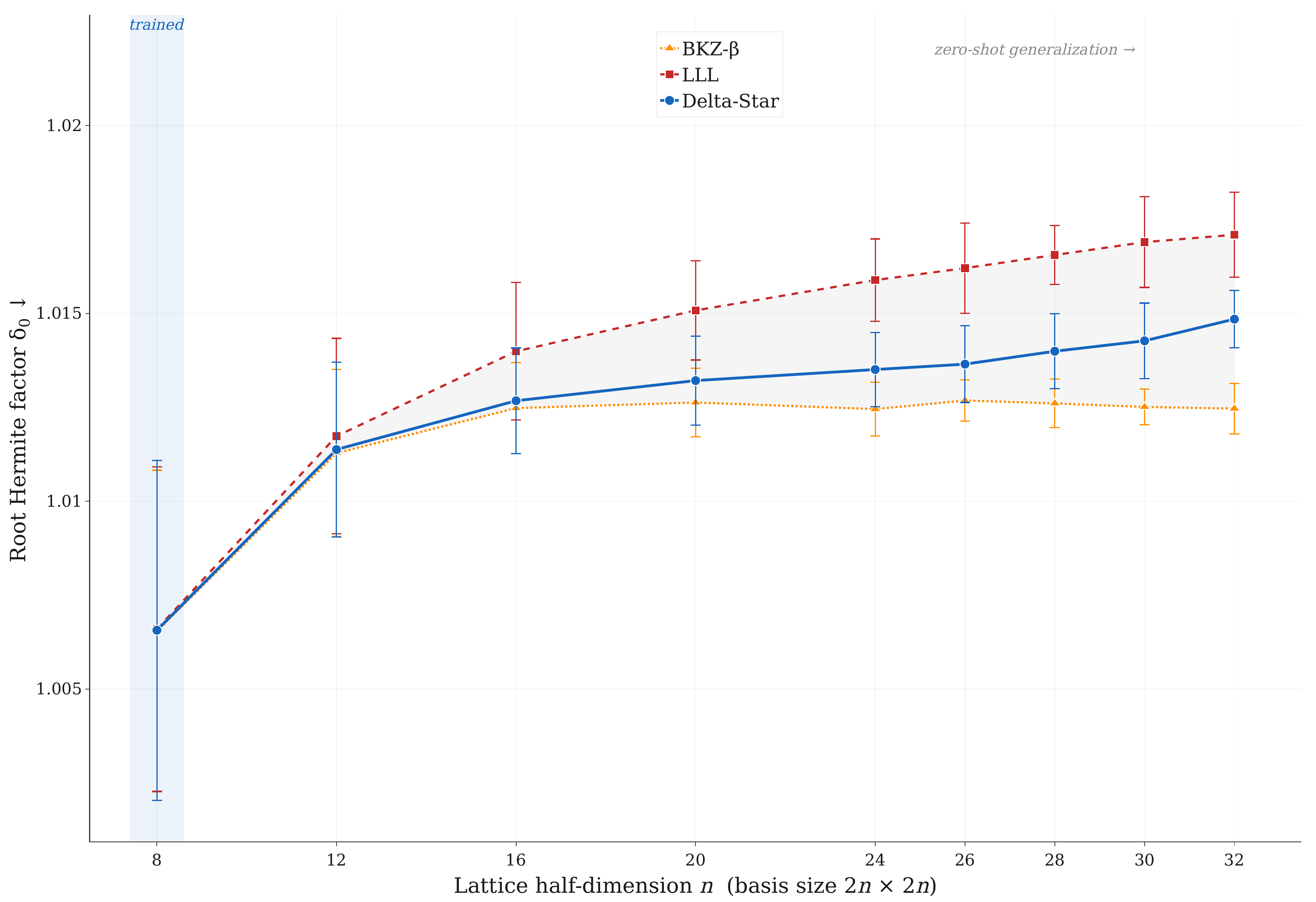}
    \end{minipage}
    \hfill
    \begin{minipage}[c]{0.54\textwidth}
        \centering
        \renewcommand{\arraystretch}{1.5} 
        \resizebox{\textwidth}{!}{%
        \begin{tabular}{ccccc}
        \hline
        $n$ & \Star{} $\delta_0$ & LLL $\delta_0$ & BKZ $\delta_0$ & $\Delta$(LLL-\Star{}) \\
        \hline
        8  & $1.00657 \pm 0.00452$ & $1.00660 \pm 0.00432$ & $1.00654 \pm 0.00428$ & 0.00003 \\
        12 & $1.01138 \pm 0.00233$ & $1.01173 \pm 0.00260$ & $1.01128 \pm 0.00223$ & 0.00036 \\
        16 & $1.01267 \pm 0.00140$ & $1.01399 \pm 0.00183$ & $1.01248 \pm 0.00121$ & 0.00132 \\
        20 & $1.01321 \pm 0.00118$ & $1.01508 \pm 0.00132$ & $1.01263 \pm 0.00092$ & 0.00187 \\
        24 & $1.01351 \pm 0.00099$ & $1.01589 \pm 0.00110$ & $1.01245 \pm 0.00071$ & 0.00238 \\
        26 & $1.01365 \pm 0.00102$ & $1.01620 \pm 0.00120$ & $1.01268 \pm 0.00055$ & 0.00256 \\
        28 & $1.01399 \pm 0.00100$ & $1.01655 \pm 0.00079$ & $1.01260 \pm 0.00065$ & 0.00256 \\
        30 & $1.01427 \pm 0.00100$ & $1.01690 \pm 0.00121$ & $1.01251 \pm 0.00048$ & 0.00263 \\
        32 & $1.01485 \pm 0.00076$ & $1.01709 \pm 0.00113$ & $1.01246 \pm 0.00067$ & 0.00225 \\
        \hline
        \end{tabular}%
        }
    \end{minipage}
    \caption{Zero-shot scaling of final basis quality across dimensions. \Star{} (blue), trained only on $n=8$, consistently outperforms LLL (red) as dimension increases, staying close to the BKZ-$\beta$ lower bound (orange) up to $n=32$. Error bars  and table show one standard deviation over 50 instances.}
    \label{fig:scaling_summary}
\end{figure}

The agent generalizes remarkably well, and its advantage over LLL grows with dimension. Figure~\ref{fig:scaling_summary} summarizes the final root Hermite factor $\delta_0$ across all dimensions from $n=8$ to $n=32$. At the training dimension ($n=8$), \Star{} converges to a slighly lower final $\delta_0 \approx 1.00657$ than LLL's $\delta_0 = 1.0066$. However, as the dimension increases, \Star{} finds increasingly better bases compared to LLL. Up to $n=32$, \Star{} stays within the 'target zone' between LLL and BKZ-$\beta$, and at $n=32$, it achieves $\delta_0 = 1.0149$ versus LLL's $1.0171$ and BKZ-$\beta$'s $1.0125$.

\subsection{Complexity Discussion}
\label{sec:efficiency}

Beyond final basis quality, the learned policy is substantially more efficient. By decoupling cursor movement from LLL's rigid sequential traversal, the agent finds a more direct path through the state space.

\begin{figure}[ht]
    \centering
    \includegraphics[width=0.9\textwidth]{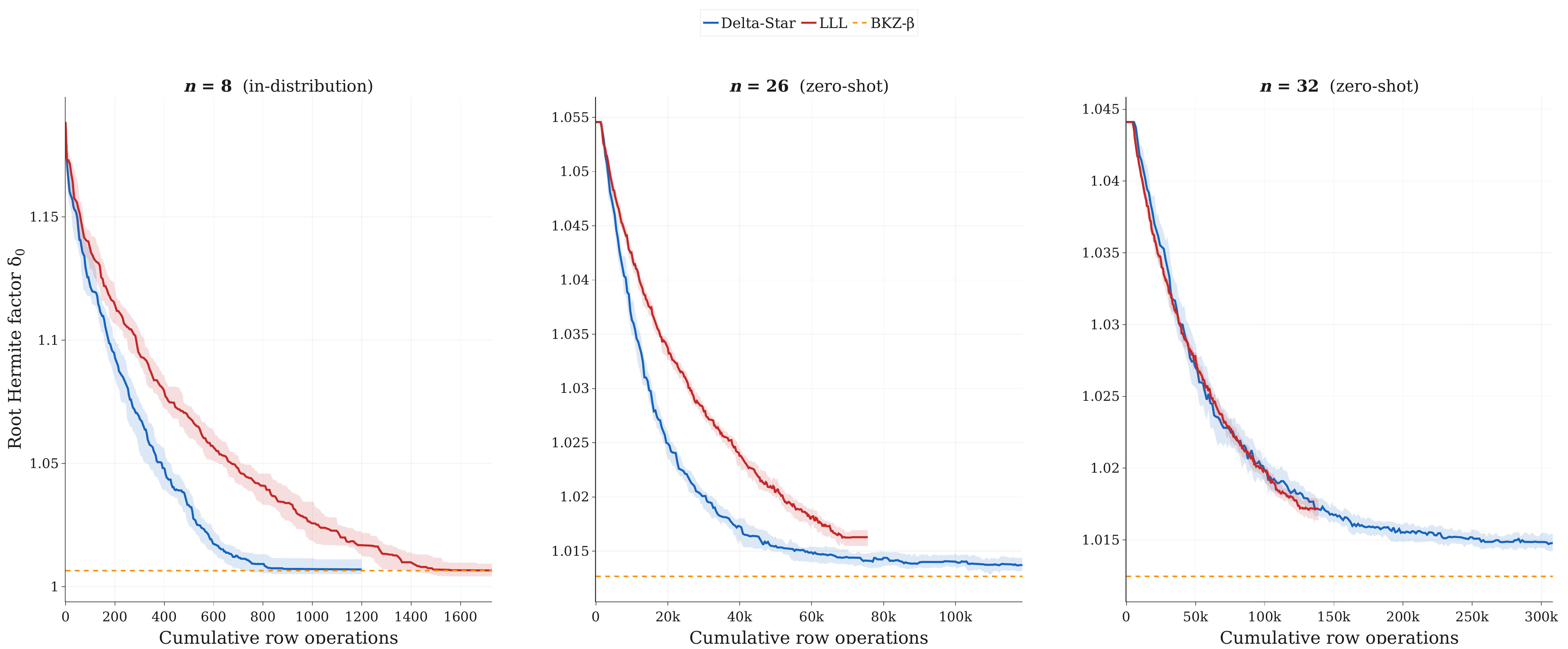}
    \caption{Reduction trajectories at $n=8$ (in-distribution) and $n=26,32$ (zero-shot generalizations). \Star{} converges faster than LLL($\delta_{LLL}=0.99$) at $n=8$ and achieves a strictly better final root Hermite factor at $n=26$ and $n=32$. Shaded regions show the interquartile range.}
    \label{fig:main_traj_scaling}
\end{figure}

Figure~\ref{fig:main_traj_scaling} contrasts the reduction trajectories at the training dimension ($n=8$) and large zero-shot dimensions ($n=26, 32$). At $n=8$, \Star{} converges to LLL's quality but requires only $55\%$ of the row operations. At $n=26$, \Star{} stays consistently below LLL's trajectory throughout the entire reduction process and converges to a strictly better final quality. Finally at $n=32$, the policy continues to reduce the bases while LLL stopped at around $\delta_0=1.017$.

\textbf{Complexity Scaling.} To characterize how the computational cost of the algorithm learned by \Star{} grows with lattice dimension, we fit scaling laws to the number of cumulative row operations the agent requires to reach its best root-Hermite factor $\delta_0$. \Cref{fig:scaling_law_ops} in the appendix plots this complexity metric as a function of the lattice dimension $2n$, with individual seed outcomes shown alongside the mean $\pm$ one standard deviation. We compare two candidate models: a polynomial $a \cdot n^{b}$ and a sub-exponential $a \cdot \exp(b\sqrt{n})$. They both reasonably fit the curve with a high coefficient of determination.

The learned policy is not competitive in wall-clock time: each action requires a neural-network forward pass, making it substantially slower per row operation than a classical implementation. Our comparison in terms of cumulative row operations therefore measures algorithmic efficiency rather than runtime performance. This distinction is deliberate. By counting elementary lattice operations, we characterize the structure of the reduction strategy the policy has discovered, independent of its current implementation complexity. Although we have not yet distilled this strategy into a closed-form algorithm, the row-operation scaling curves provide an empirical estimate of the performance and complexity such an algorithm would exhibit —  allowing us to study a classical algorithm's properties before having derived it.


\section{Related Work}
\label{sec:related}
\subsection{Classical Lattice Reduction}

The foundation of polynomial-time lattice reduction was established by the LLL algorithm \cite{lenstra1982factoring}. Schnorr introduced stronger reduction hierarchies \cite{BKZ}, culminating in the Block Korkine-Zolotarev (BKZ) algorithm \cite{schnorr1994lattice, li2025bkz}. BKZ generalizes the Lov\'asz condition to block size $\beta \ge 2$, requiring exact SVP solutions in projected sublattices, which yields shorter vectors at exponential complexity in $\beta$.

Recent work has focused on computational efficiency. Nguyen and Stehl{\'e} \cite{nguyen2009l2} introduced the $L^2$ algorithm, a provably polynomial-time floating-point LLL variant forming the basis of modern implementations. Novocin et al.\ \cite{novocin2011l1} achieved the first quasi-linear (in bit-length) time complexity for LLL via a recursive structure. Building on this, Flatter \cite{flatter} speeds up reduction through recursive precision and compression, while BLASter \cite{ducas2025modern} optimizes performance using BLAS operations. These methods target speed, outputting standard LLL-reduced bases. Our work instead aims to improve reduction quality (producing shorter vectors) strictly within LLL's original, local action space. This task's difficulty is underscored by Micciancio's result that approximating the shortest vector to within constant factors is NP-hard \cite{micciancio2001complexity}.

\subsection{Machine Learning for Combinatorial Optimization}

There is a growing literature on applying machine learning to combinatorial optimization \cite{bengio2021machine, khalil2017learning, bello2017neural, zhou2025urs}. Our approach builds on the integration of deep neural networks with MCTS \cite{coulom2006efficient, browne2012survey}, pioneered by AlphaGo \cite{silver2016mastering} and generalized by AlphaZero \cite{silver2018general}. Since lattice reduction lacks an adversary, we adapt MCTS for single-player environments, following prior work on puzzles \cite{schadd2012single, dantsin2022alphazero}.

Our work is most closely related to the recent line of AI-driven algorithm discovery: AlphaTensor \cite{fawzi2022discovering} for matrix multiplication, AlphaDev \cite{mankowitz2023alphadev} for sorting, FunSearch \cite{romera2024funsearch} for combinatorial structures, AlphaEvolve \cite{novikov2025alphaevolve} for general coding tasks, AlphaProof \cite{hubert2025alphaproof} and AlphaGeometry \cite{trinh2024alphageometry} for mathematical reasoning. We extend this paradigm to lattice reduction.

\subsection{Machine Learning for Lattice Problems}

Most prior work at the intersection of machine learning and lattice problems targets cryptanalysis rather than algorithm discovery. The SALSA series \cite{salsa, picante, verde, coolcruel, fresca} uses transformer architectures \cite{transformer17} to attack LWE instances by recovering sparse secrets.

Two prior works have directly applied machine learning to lattice reduction. Marchetti et al.\ \cite{neurallattice2023} proposed a self-supervised geometric deep learning approach that directly outputs a factorized unimodular transformation matrix. Their model is invariant to isometries and scaling but achieves performance comparable to, not exceeding, LLL up to dimension $8$. Cheong et al.\ \cite{cheong2025bkz} used deep RL to dynamically select the BKZ block size $\beta$ via a transformer-based model. While their agent discovered non-trivial strategies (such as prioritizing mid-range block sizes), performance plateaued during training and the model failed to generalize to higher dimensions.

Our approach differs from both works in three respects. First, we optimize the core reduction decisions (swaps and size reductions) within LLL's exact action space, rather than predicting full transformations or selecting hyperparameters for BKZ. Second, we employ a ResNet within an AlphaZero-style self-play loop augmented with single-player adaptive horizon MCTS (Section~\ref{sec:adaptive_mcts}). Third, our model strictly outperforms LLL in reduction quality and generalizes zero-shot to dimensions well beyond those seen during training.

\subsection{Adaptive Horizon and Entropy in MCTS}

Our entropy-gated adaptive horizon MCTS relates to several efforts to deepen search efficiently. V-MCTS \cite{ye2022vmcts} uses multi-step virtual expansions to approximate final policies for early termination, while 3HNN \cite{gao2018threehead} introduces an action-value head to utilize neural evaluations even when node expansion is delayed. Other works incorporate entropy into MCTS, such as ANTS \cite{kozakowski2021ants} and EG-MCTS \cite{li2024entropy}, but primarily to bias action selection rather than control expansion depth. While MuZero \cite{schrittwieser2020muzero} and EfficientZero \cite{ye2021efficientzero} use multi-step unrolling during training, their search expansion remains standard. The predictive power of MCTS plans has been noted for interpretability \cite{chung2024predicting}, and macro-actions \cite{dewaard2016options, hafner2022director} allow multi-step commitments, but these typically rely on pre-defined options or learned abstractions rather than online policy confidence. Finally, while MCTS has been applied to non-game optimization like resource allocation \cite{bertsimas2017mcts} and financial hedging \cite{szehr2023hedging}---sometimes producing asymmetric trees via standard UCB allocation---\Star{} is unique in explicitly using the Shannon entropy of the policy network to dynamically gate the physical expansion depth within a single iteration, skipping forced sequences to concentrate compute on genuine decision points.

\section{Conclusion}
\label{sec:conclusion}
We have demonstrated that deep reinforcement learning can discover lattice reduction strategies that are strictly superior to the classical LLL algorithm. By formulating reduction as a Markov Decision Process over primitive row operations and training a value-policy network via self-play, our model, \Star{}, learns a scale-invariant heuristic. Despite being trained exclusively on small $8$-dimensional $q$-ary lattices, \Star{} generalizes zero-shot to dimensions up to $n=30$ and unseen moduli, consistently converging in significantly fewer operations than LLL while achieving better basis quality that approaches BKZ.

This work opens several compelling directions for future research. First, the learned policy weights encode a deterministic, highly efficient reduction algorithm; extracting and formalizing this classical algorithm from the neural network is a primary objective. Second, the same self-play methodology can be applied to improve block-based reduction algorithms like BKZ by augmenting the action space with oracle calls for exact Shortest Vector Problem (SVP) subroutines. Finally, the success of \Star{} suggests that other fundamental, heuristic-driven algorithms in computational mathematics and cryptography may be similarly accelerated by discovering optimal decision paths through reinforcement learning.

\bibliographystyle{plainnat}
\bibliography{references}

\appendix

\section{Additional Lattice Background}
\label{sec:appendix_lattice_background}

\subsection{Orthogonality Defect and Its Relationship to the Root Hermite Factor}

The orthogonality defect of a basis $\mathbf{B}$ is defined as:
\begin{equation}
    \delta(\mathbf{B}) = \frac{\prod_{i=1}^n \|\mathbf{b}_i\|}{\det(\mathcal{L})} = \prod_{i=1}^n \frac{\|\mathbf{b}_i\|}{\|\widetilde{\mathbf{b}}_i\|}.
\end{equation}
Hadamard's inequality guarantees $\delta(\mathbf{B}) \ge 1$, with equality if and only if the basis is orthogonal. A globally orthogonalized basis naturally forces individual vectors to be shorter, so minimizing the orthogonality defect tends to improve the root Hermite factor as well. In practice, lattice reduction algorithms seek a unimodular transformation $\mathbf{U}$ that simultaneously reduces both quantities.

\subsection{The LLL Potential Function}

The standard termination and complexity analysis of LLL relies on the potential function:
\begin{equation}
    \Phi(\mathbf{B}) = \prod_{i=1}^{n} \|\widetilde{\mathbf{b}}_i\|^{n-i+1}.
\end{equation}
Every swap operation strictly decreases $\Phi$ by a factor of at least $\sqrt{\delta} < 1$, while size-reduction operations leave $\Phi$ unchanged. Because $\Phi$ is bounded below by a function of $\det(\mathcal{L})$, the total number of swaps is polynomial. The Lov\'asz condition is designed precisely to guarantee this monotonic decrease: it is the sufficient condition under which a swap provably reduces the potential. However, the condition does not prescribe the optimal sequence of actions for achieving the highest quality reduction.

\section{Ablation Study: Reward Model Selection and Benchmark}
\label{sec:appendix_reward_ablation}

A critical design choice in our MDP formulation is the reward function. As introduced in Section~\ref{sec:game}, the hybrid reward is parameterized by a potential weight $p \in [0, 1]$, interpolating between the orthogonality defect ($p=0$) and the LLL potential function ($p=1$).

We trained five independent models for $p \in \{0.0, 0.25, 0.5, 0.75, 1.0\}$ on $n=8$ lattices up to 100,000 training steps, then evaluated the best checkpoint for each model across 100 random instances per condition. The benchmark varies three parameters: lattice modulus $q$, dimension $n$, and maximum step allocation $T_{\max}$. All models are evaluated on the root Hermite factor $\delta_0$, a neutral metric independent of the training objective.

\subsection{In-Distribution Convergence and Efficiency}

\begin{figure}[H]
    \centering
    \includegraphics[width=\textwidth]{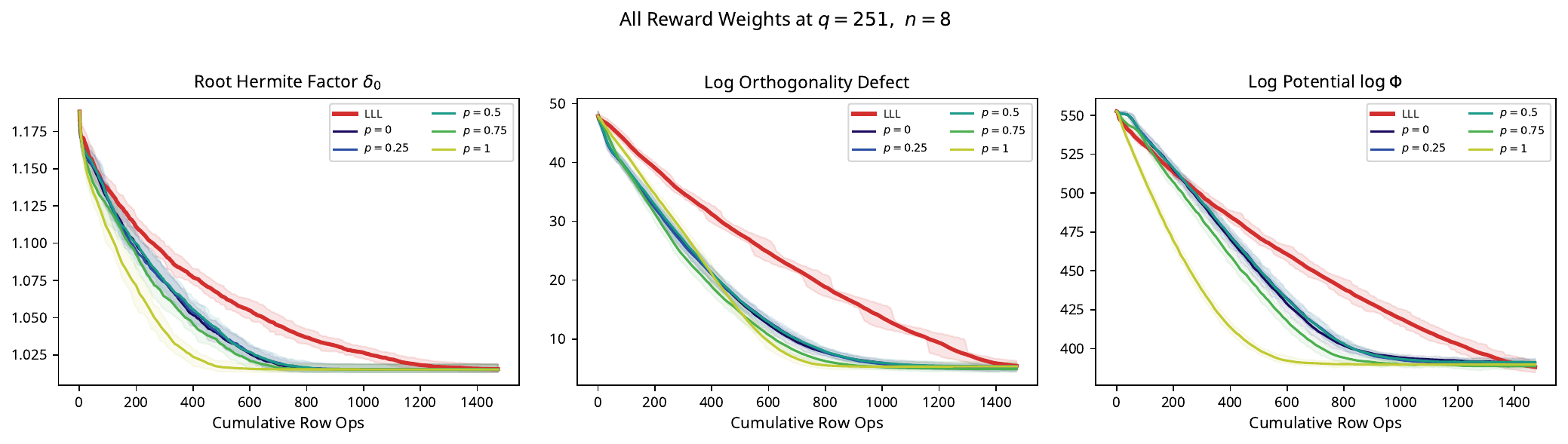}
    \caption{Trajectories for all five reward models at $q=251$. All converge to similar basis quality, but $p=1$ (yellow-green) converges fastest.}
    \label{fig:app_allpotw_q251}
\end{figure}

Figure~\ref{fig:app_allpotw_q251} shows the reduction trajectories for all five models at the in-distribution parameters ($n=8, q=251$). All values of $p$ converge to nearly identical final $\delta_0 \approx 1.0153$, matching LLL. The models differ in convergence speed.

When measuring the ratio of row operations required by each RL agent relative to LLL to reach $95\%$ of LLL's total $\delta_0$ improvement, every RL formulation requires significantly fewer operations. The $p=1$ model is the most aggressive, requiring only $\sim 36$--$40\%$ of LLL's operations for $q \ge 97$. However, $p=0.75$ offers the best balance of efficiency ($\sim 40$--$65\%$ of LLL's operations across all tested dimensions) and stable zero-shot generalization, which is why it was selected for the main experiments in Section~\ref{sec:experiments}.

\subsection{Generalization Across Moduli and Dimensions}

We evaluate the learned policies on unseen moduli $q \in \{23, 97, 251, 499, 1009, 4019, 10007\}$ and dimensions up to $n=20$.

\begin{figure}[ht]
    \centering
    \includegraphics[width=\textwidth]{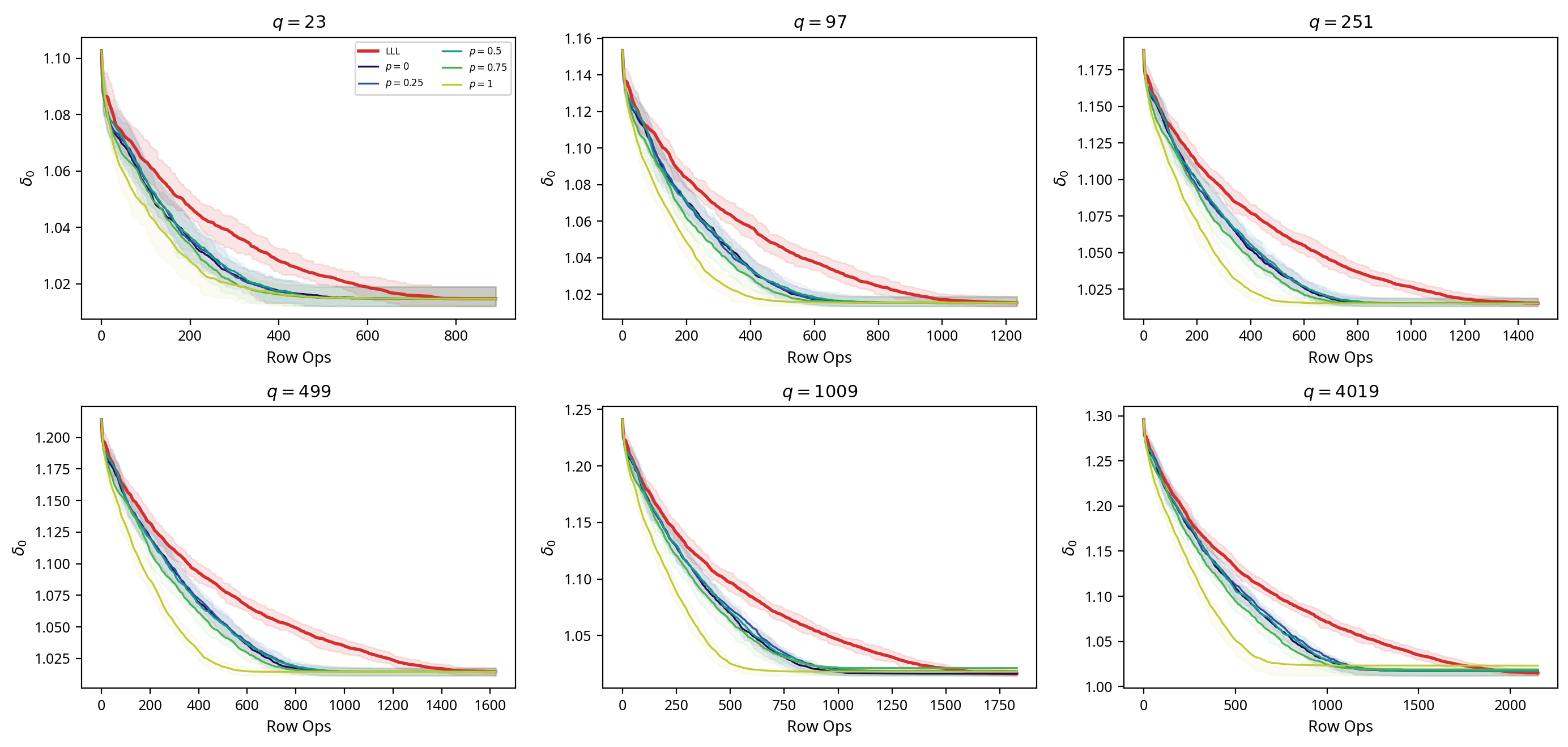}
    \caption{Root Hermite factor trajectories across lattice moduli $q$. The x-axis is clipped to LLL's cumulative row operations for each $q$. Hybrid models generalize better to larger moduli.}
    \label{fig:app_grid_q_rhf}
\end{figure}

Figure~\ref{fig:app_grid_q_rhf} shows $\delta_0$ trajectories across moduli. All models converge to LLL-quality bases at moderate moduli ($q \le 499$). At extreme out-of-distribution moduli ($q \ge 1009$), the pure orthogonality defect model ($p=0$) and hybrid models ($p=0.5, 0.75$) degrade more gracefully than $p=1$.

\begin{figure}[ht]
    \centering
    \includegraphics[width=0.9\textwidth]{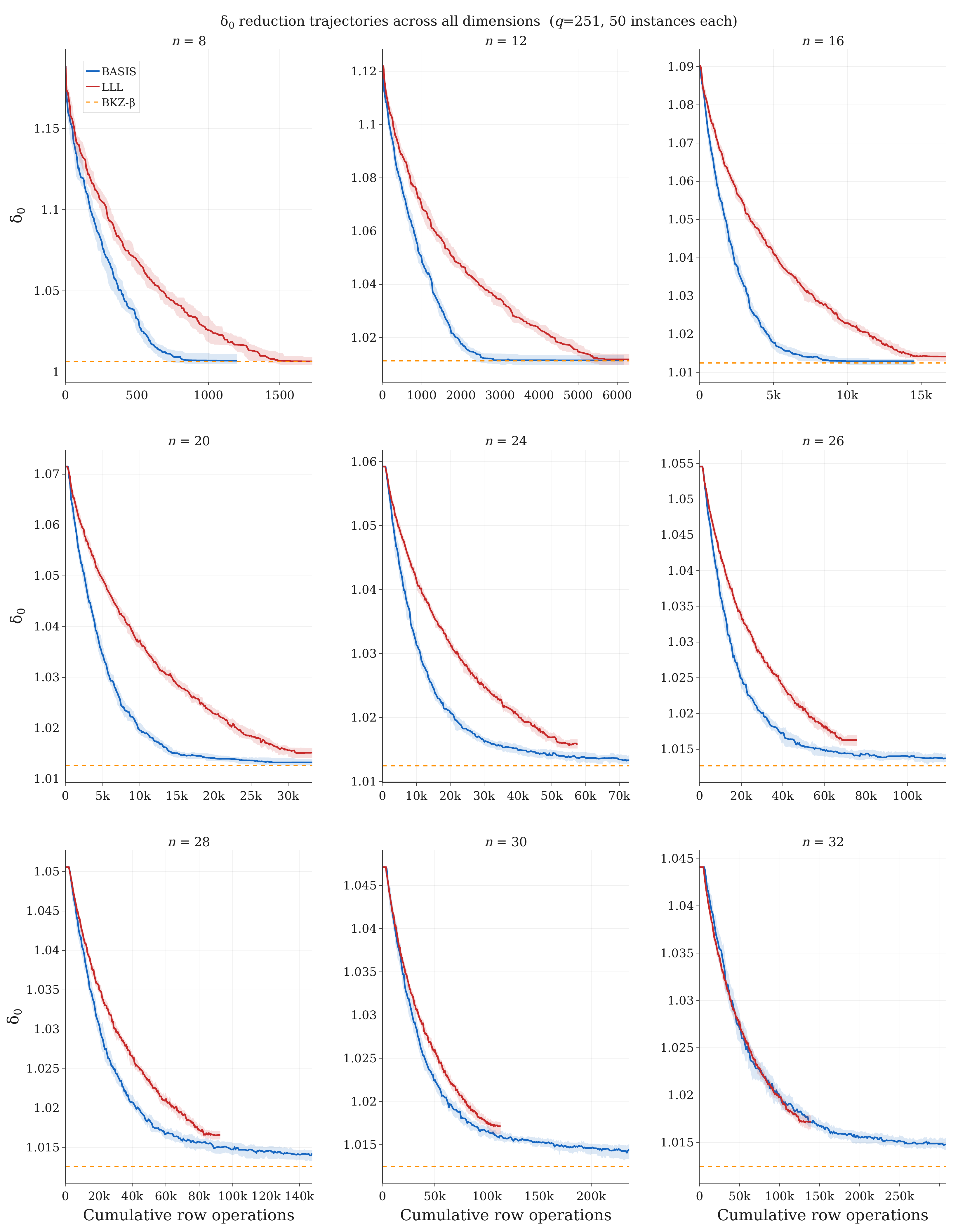}
    \caption{Root Hermite factor trajectories across all tested dimensions $n \in \{8, 12, 16, 20, 24, 26, 28, 30, 32\}$. The convergence speed advantage is preserved at larger, unseen dimensions, up to 32 where the policy continues to reduce the basis while LLL (with $\delta=0.99$) stopped at a high rhf value. The policy starts to diverge too from the BKZ performance.}
    \label{fig:app_grid_n}
\end{figure}

Generalization across dimension is the most notable property of the learned policies. As detailed in Section~\ref{sec:experiments}, the $p=0.75$ model scales zero-shot to $n=30$, consistently outperforming LLL. Figure~\ref{fig:app_grid_n} shows trajectories for the root Hermite factor across all eight tested dimensions from $n=8$ to $n=30$. The RL agent consistently converges faster than LLL at every dimension, confirming that the network has learned a scale-invariant reduction heuristic. At dimensions $n \ge 24$, the agent's trajectory stays strictly below LLL's throughout the entire reduction process.

\begin{figure}[ht]
    \centering
    \includegraphics[width=\textwidth]{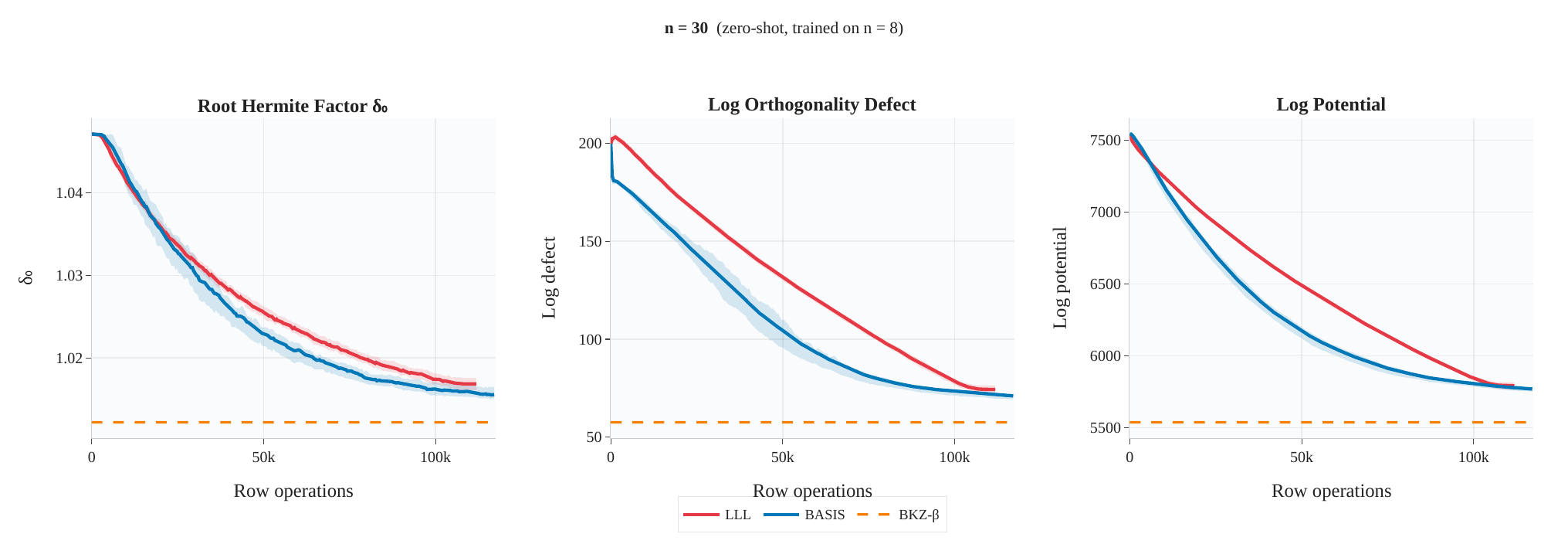}
    \caption{Detailed reduction trajectories at $n=30$ (zero-shot) across all three metrics. \Star{} (blue) converges faster and to a better final quality than LLL (red) on root Hermite factor, log orthogonality defect, and log potential.}
    \label{fig:traj_n30}
\end{figure}

To provide a complete picture of the agent's behavior at the largest tested dimension, Figure~\ref{fig:traj_n30} shows the reduction trajectories at $n=30$ across all three tracked metrics: root Hermite factor, log orthogonality defect, and log potential. \Star{} dominates LLL on all three metrics, demonstrating that the learned policy optimizes the global basis structure more effectively than the classical heuristic.

\subsection{Convergence Under Varying Allocations}

\begin{figure}[ht]
    \centering
    \includegraphics[width=\textwidth]{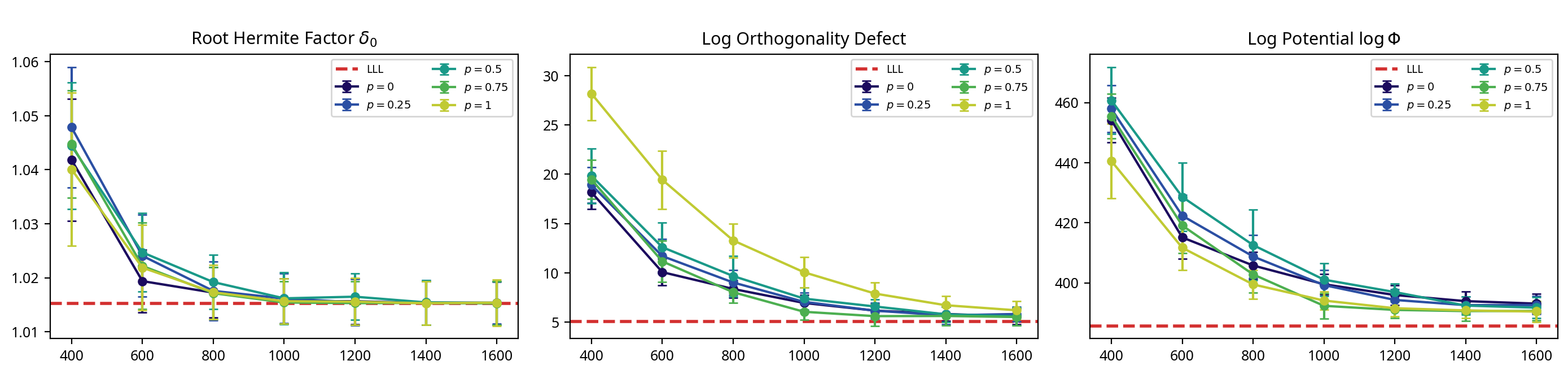}
    \caption{Final basis quality as a function of $T_{\max}$. The RL agents smoothly improve with increased step allocation, acting as anytime algorithms.}
    \label{fig:app_budget}
\end{figure}

Figure~\ref{fig:app_budget} shows performance under constrained step allocations. LLL requires a fixed number of operations and cannot be halted early without severe quality penalties. The RL agents behave as anytime algorithms: basis quality improves smoothly as the allocation increases, converging to LLL-equivalent quality by $T_{\max} \approx 1400$.

The choice of $p$ dictates a trade-off between convergence speed and generalization robustness. The pure potential formulation ($p=1$) converges fastest in-distribution. For zero-shot generalization to different dimensions and moduli, $p=0.75$ provides the best balance, enabling the scaling results in the main text.

\section{Ablation Study: Entropy-Gated Horizon}
\label{sec:appendix_horizon_ablation}

The entropy-gated adaptive horizon mechanism introduced in Section~\ref{sec:adaptive_mcts} controls the depth of multi-step expansion during MCTS. Each simulation consumes exactly one neural network evaluation regardless of horizon depth: the multi-step expansion is computationally free, amortizing a single forward pass across multiple node creations. We conduct a preliminary ablation at dimension $n=12$, evaluating 11 runs (1--2 trials per configuration) across the baseline ($H=1$) and multi-step configurations ($H \in \{4, 8\}$) with entropy thresholds $\tau \in \{0.6, 1.0, 1.5\}$. All agents use 25 MCTS simulations per move and are trained for 25--32 million environment samples.

\subsection{Performance}

Table~\ref{tab:horizon_summary} summarizes the results. All horizon variants outperform the baseline, suggesting that multi-step expansion improves learning when the simulation budget is held constant. The best-performing configurations are $H=4, \tau=1.5$ and $H=8, \tau=0.6$, which achieve comparable returns despite operating in very different regimes.

\begin{table}[ht]
\centering
\caption{Horizon ablation at $n=12$. Metrics are averaged over the final quarter of training. Total nodes estimates the tree size per action decision ($25 \times$ horizon depth). Results should be interpreted cautiously given the limited number of trials per configuration.}
\label{tab:horizon_summary}
\begin{tabular}{lcccccc}
\toprule
Config & Horizon Depth & Sel.\ Depth & Total Depth & $\approx$ Nodes & Return \\
\midrule
$H=1$ (baseline) & 1.00 & 26.4 & 27.4 & 25 & $-0.211$ \\
$H=4,\;\tau=0.6$ & 2.69 & 38.2 & 40.8 & 67 & $-0.207$ \\
$H=4,\;\tau=1.0$ & 3.25 & 33.8 & 37.1 & 81 & $-0.197$ \\
$H=4,\;\tau=1.5$ & 3.89 & 35.2 & 39.1 & 97 & $-0.196$ \\
$H=8,\;\tau=0.6$ & 3.85 & 40.8 & 44.6 & 96 & $-0.197$ \\
$H=8,\;\tau=1.0$ & 4.76 & 29.5 & 34.2 & 119 & $-0.201$ \\
\bottomrule
\end{tabular}
\end{table}

The data hints at an interaction between horizon capacity $H$ and the entropy threshold $\tau$. At $H=4$, higher thresholds appear beneficial ($\tau=1.5 > 1.0 > 0.6$): with bounded capacity, the maximum commitment of 4 steps may be short enough that errors remain recoverable. At $H=8$, the relationship appears to invert: $\tau=0.6$ outperforms $\tau=1.0$. When the horizon allows very deep commitments, aggressive expansion risks committing to sequences that are not truly forced, potentially injecting unreliable value estimates into the tree. Strict gating prevents this by restricting expansion to high-confidence states. We note, however, that with at most 2 trials per configuration, these trends remain preliminary.

\subsection{Self-Regulating Dynamics}

\begin{figure}[ht]
    \centering
    \includegraphics[width=\textwidth]{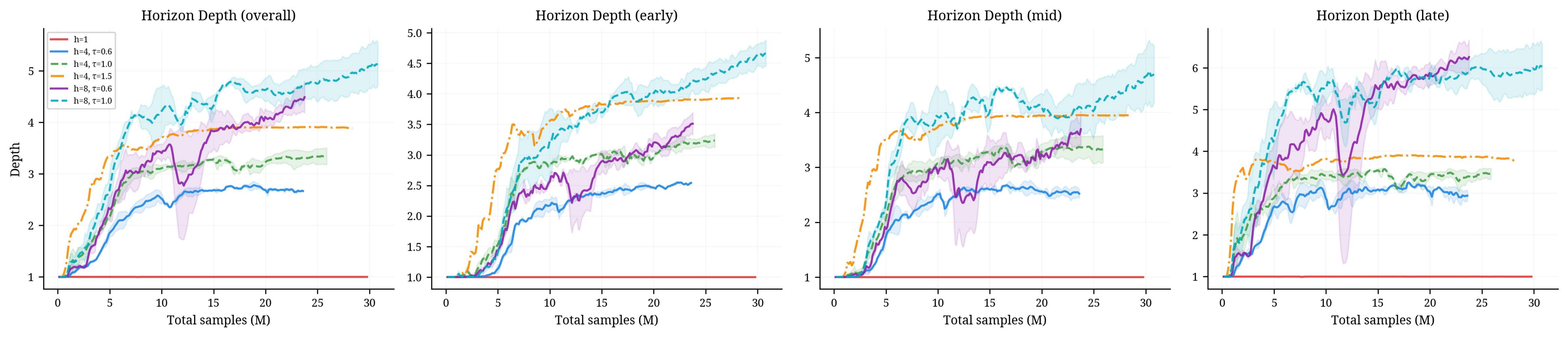}
    \caption{Horizon depth over the course of training. Strict-gating configurations ($\tau=0.6$) start near the baseline and grow as the policy gains confidence, while high-threshold configurations start high and plateau.}
    \label{fig:app_horizon_depth}
\end{figure}

The horizon mechanism appears to exhibit a self-regulating feedback loop during training. As shown in Figure~\ref{fig:app_horizon_depth}, the average horizon depth for $H=8, \tau=0.6$ grows from 1.69 in the first quarter of training to 4.29 in the final quarter---a 153\% increase with no sign of saturation. This growth is not externally scheduled: it emerges as the policy becomes more confident (lower entropy), allowing more states to pass the entropy gate.

Across all horizon variants, the temporal correlation between horizon depth and return is positive ($r \approx +0.88$ to $+0.96$), while the correlation between policy entropy and return is negative ($r \approx -0.93$). This is consistent with a virtuous cycle: as the policy improves, entropy decreases, more states pass the gate, and the tree captures longer forced sequences. Multi-step targets also increase gradient norms (up to $2\times$ the baseline), but all runs remain stable throughout training.

\section{Zero-Shot Generalization to Ring-LWE and Module Lattices}
\label{sec:appendix_rlwe}

In addition to standard unstructured $q$-ary lattices, we evaluate our agent on structured ideal lattices underlying the Ring Learning With Errors (Ring-LWE) problem \cite{Lyubashevskyetal2010}. Ring-LWE forms the mathematical foundation of several NIST post-quantum cryptographic standards.

\subsection{Kannan's Embedding for Primal Attacks}

The Ring-LWE problem asks an adversary to recover a secret polynomial $\mathbf{s} \in \mathcal{R}_q$ from noisy samples $(\mathbf{a}, \mathbf{b} = \mathbf{a} \cdot \mathbf{s} + \mathbf{e} \pmod q)$, where $\mathcal{R}_q = \mathbb{Z}_q[x]/(x^n + 1)$ and the coefficients of $\mathbf{s}$ and $\mathbf{e}$ are drawn from a narrow centered binomial distribution parameterized by $\eta$.

The standard primal attack frames secret recovery as finding a uniquely short vector in a specific lattice via Kannan's embedding. For a single Ring-LWE sample, we construct a $(2n+1) \times (2n+1)$ basis:
\begin{equation}
    \mathbf{B}_{\text{RLWE}} = \begin{bmatrix}
    q\mathbf{I}_n & \mathbf{0} & \mathbf{0} \\
    \mathbf{A}_{\text{neg}} & \mathbf{I}_n & \mathbf{0} \\
    \mathbf{b}^\top & \mathbf{0} & c
    \end{bmatrix},
\end{equation}
where $\mathbf{A}_{\text{neg}}$ is the $n \times n$ negacyclic matrix representing polynomial multiplication by $\mathbf{a}$ in $\mathcal{R}_q$, and $c$ is a scalar embedding factor (typically $c=1$).

This lattice contains the target vector $\mathbf{v} = (\mathbf{e}, -\mathbf{s}, c)$. Because $\mathbf{s}$ and $\mathbf{e}$ have small coefficients, $\mathbf{v}$ is unusually short relative to the expected shortest vector in a random lattice of the same volume. A successful lattice reduction will place $\mathbf{v}$ among the rows of the reduced basis, allowing direct recovery of the secret key.

\subsection{Secret Recovery}

We train and evaluate our agent on Ring-LWE instances. The environment state includes the $(2n+1) \times (2n+1)$ basis, and the reward function includes a terminal bonus for recovering the target vector $(\mathbf{e}, -\mathbf{s}, c)$.

The agent generalizes zero-shot to larger dimensions in this domain as well. Using the same primitive action space (\textsc{MoveUp}, \textsc{MoveDown}, \textsc{Swap}, \textsc{SizeReduce}), the learned policy reduces the Ring-LWE lattice and recovers the secret polynomial.

\section{Figures}

\begin{figure}[ht]
    \centering
    \includegraphics[width=1.0\linewidth]{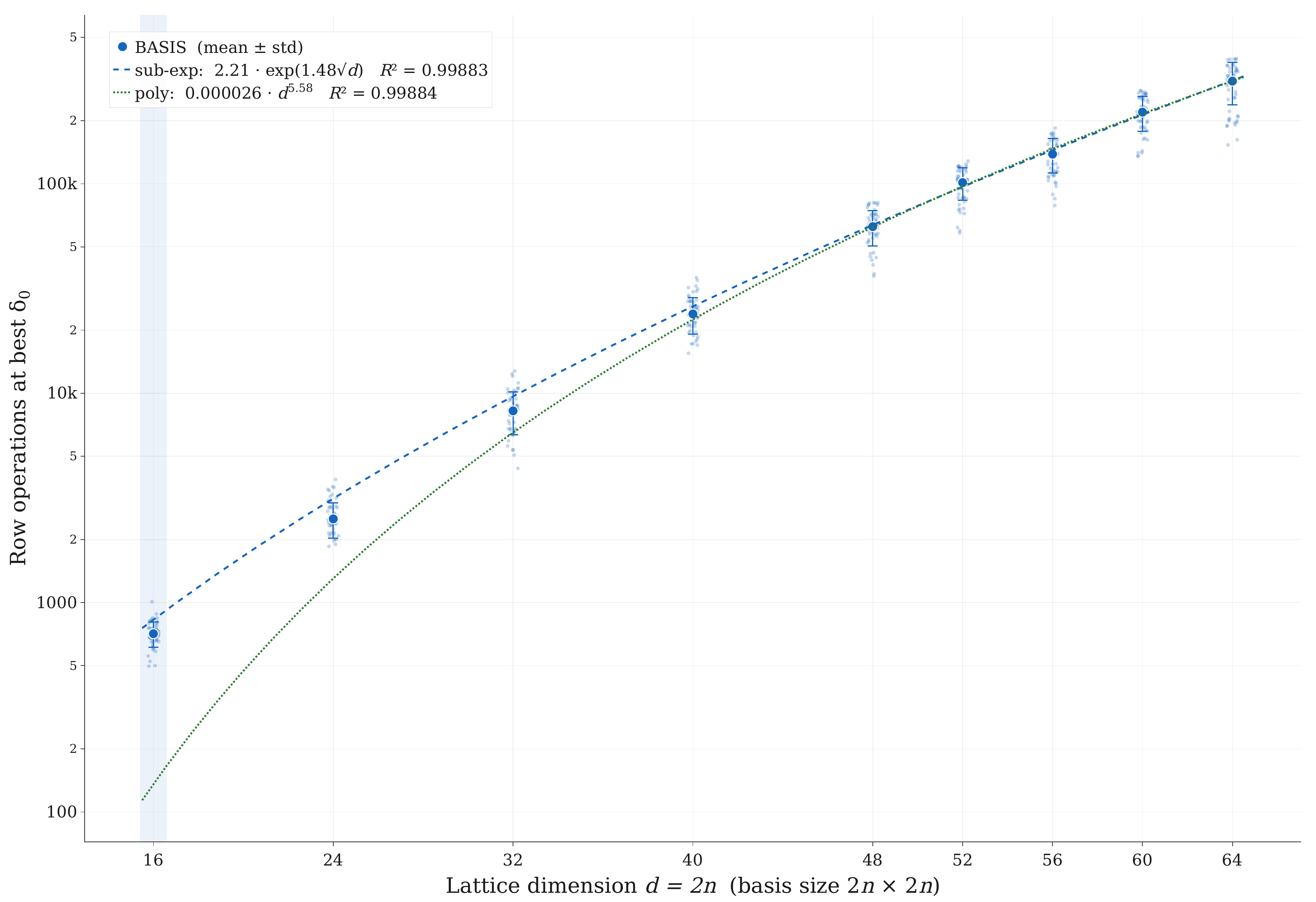}
    \caption{Complexity scaling of \Star{}. Number of cumulative row operations required to reach the best root-Hermite factor $\delta_0$ as a function of the lattice dimension $d = 2n$. Blue circles show the mean across seeds with error bars indicating $\pm$ one standard deviation; translucent dots show individual seed outcomes. The shaded region marks the training dimension ($d{=}16$). Two scaling models are fit to the empirical means: a sub-exponential $a \cdot \exp(b\sqrt{d})$ (blue dashed; $a{=}2.21$, $b{=}1.48$, $R^2{=}0.999$) and a polynomial $a \cdot d^{b}$ (green dotted; $b{=}5.58$, $R^2{=}0.999$). Both models achieve comparable coefficients of determination.}
    \label{fig:scaling_law_ops}
\end{figure}

\subsection{Analyzing Policy Traces}

To characterize the learned strategy beyond aggregate statistics, we
mined the top-$k$ longest contiguous action subsequences shared across
all 100 greedy-policy trajectories using a generalized suffix-array
algorithm.  The results reveal a strikingly regular two-phase cycle.
Every trajectory is composed of alternating
\emph{descent} and \emph{cascade} phases.  During a descent phase the
policy moves the cursor downward through the basis, interleaving
\textsc{MoveDown} and \textsc{SizeReduce} actions in an approximate
$(\textsc{D}\,\textsc{R})^{m}$ pattern; during a cascade phase it
executes a burst of consecutive \textsc{Swap} operations
($\textsc{W}^{n}$) that bubble the current vector toward the top of
the basis.  The longest motif present in all 100 trajectories is
$\textsc{D}\,\textsc{R}\,\textsc{D}\,\textsc{R}\,\textsc{W}^{12}$
(length 16), capturing the canonical
descent-to-cascade transition.  The seven longest motifs (lengths
12--16) all straddle this phase boundary, confirming that the
transition point is the most stereotyped part of the policy.  Shorter
motifs ($\ell \le 10$) include the pure-descent pattern
$(\textsc{D}\,\textsc{R})^{5}$ (756 occurrences) and the pure-descent
run $\textsc{D}^{7}$ (942 occurrences), the latter corresponding to
rapid cursor repositioning without size reduction.  The learned
algorithm thus resembles an aggressive deep-insertion strategy, where
the policy first descends to identify a promising vector near the
bottom of the basis, then commits to a full swap cascade that inserts
it near the top --- a pattern closer to deep insertion
LLL~\cite{schnorr1994lattice} than to the standard algorithm, but
discovered autonomously by reinforcement learning.

\begin{figure}[ht]
    \centering
    \includegraphics[width=0.9\textwidth]{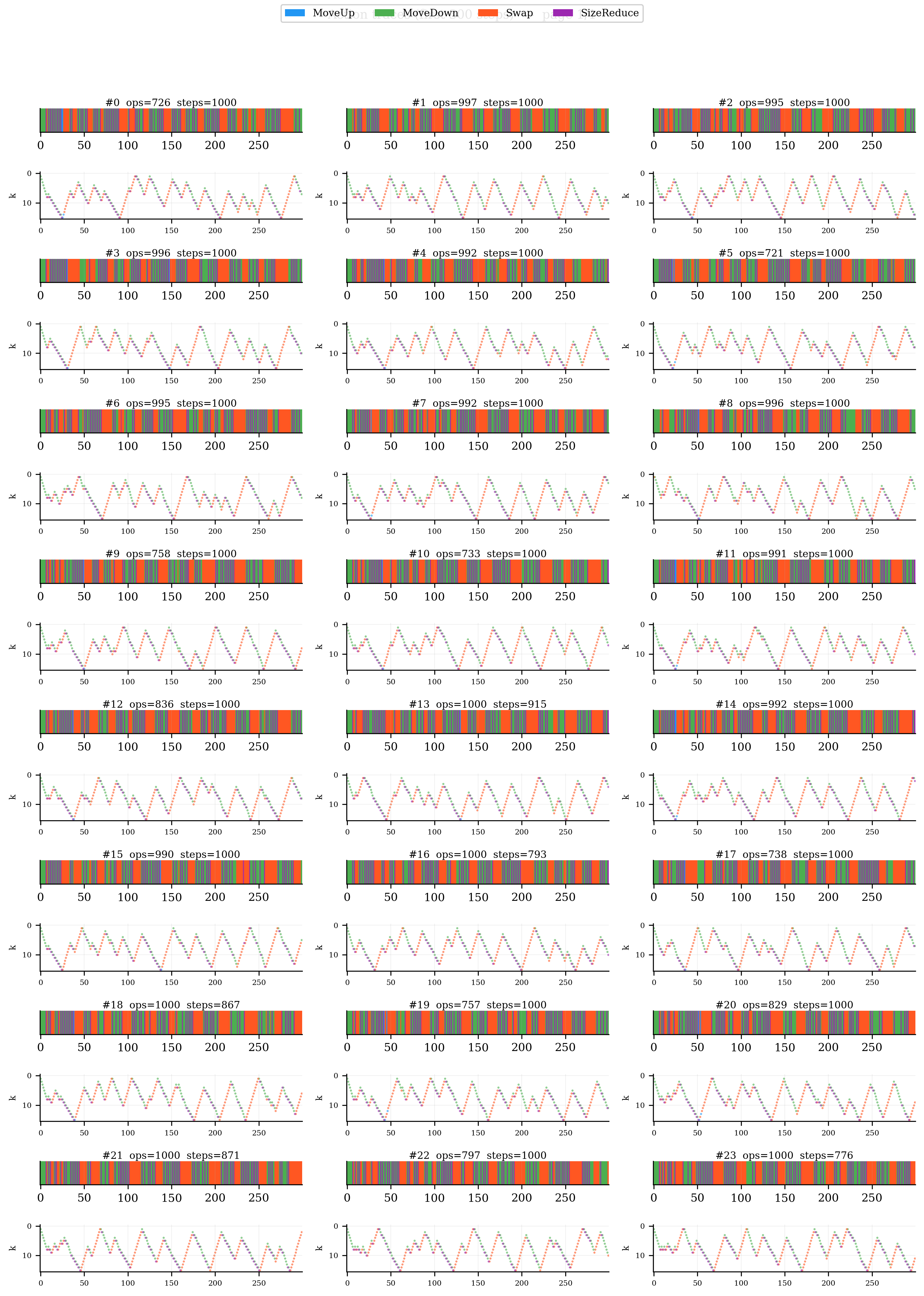}
    \caption{Action traces of the learned policy over the first 300 environment steps across multiple q-ary lattice instances ($n=8$, $q=251$). Each panel shows two rows per instance: a colour-coded action ribbon (\textcolor{blue!70}{MoveUp}, \textcolor{green!60!black}{MoveDown}, \textcolor{red!70}{Swap}, \textcolor{violet}{SizeReduce}) and the cursor position~$k$ before each action. The policy exhibits a consistent sweep--reduce--cascade pattern across diverse random seeds.}
    \label{fig:action-traces}
\end{figure}

\clearpage

\end{document}